\documentclass[10pt,twocolumn,letterpaper]{article}
\usepackage[pagenumbers]{cvpr}

\usepackage{graphicx}
\usepackage{amsmath}
\usepackage{amssymb}
\usepackage{booktabs}
\usepackage[pagebackref,breaklinks,colorlinks]{hyperref}

\usepackage{hyperref}      
\usepackage{url}           
\usepackage{booktabs}      
\usepackage{amsfonts}      
\usepackage{nicefrac}      
\usepackage{microtype}     
\usepackage{xcolor}        
\usepackage{bbding}

\usepackage{colortbl}
\usepackage{adjustbox}
\usepackage{tabularx}
\usepackage{multirow}
\usepackage{array}
\usepackage{enumitem}
\usepackage{multicol}

\newcommand{\mypm}{\tiny$\pm$}

\usepackage[capitalize]{cleveref}
\crefname{section}{Sec.}{Secs.}
\Crefname{section}{Section}{Sections}
\Crefname{table}{Table}{Tables}
\crefname{table}{Tab.}{Tabs.}

\begin{document}

\title{StyleAdv: Meta Style Adversarial Training for Cross-Domain Few-Shot Learning}
\author{Yuqian Fu$^{1}$, Yu Xie$^{2}$, Yanwei Fu$^{3}$, Yu-Gang Jiang$^{1}\thanks{\ indicates corresponding author}$ 
\\$^1$Shanghai Key Lab of Intelligent Information Processing, School of Computer Science, Fudan University
\\$^2$Purple Mountain Laboratories, Nanjing, China. $^3$School of Data Science, Fudan University
\\ \texttt{\{fuyq20, yxie18, yanweifu, ygj\}@fudan.edu.cn}
}
\maketitle

\begin{abstract}
Cross-Domain Few-Shot Learning (CD-FSL) is a recently emerging task that tackles few-shot learning across different domains. It aims at transferring prior knowledge learned on the source dataset to novel target datasets. The CD-FSL task is especially challenged by the huge domain gap between different datasets. Critically, such a domain gap actually comes from the changes of visual styles, and wave-SAN~\cite{fu2022wave} empirically shows that spanning the style distribution of the source data helps alleviate this issue.
However, wave-SAN simply swaps styles of two images. Such a vanilla operation makes the generated styles ``real'' and ``easy'', which still fall into the original set of the source styles.
Thus, inspired by vanilla adversarial learning, a novel model-agnostic meta Style Adversarial training (StyleAdv) method together with a novel style adversarial attack method is proposed for CD-FSL. 
Particularly, our style attack method synthesizes both ``virtual'' and ``hard'' adversarial styles for model training. This is achieved by perturbing the original style with the signed style gradients.
By continually attacking styles and forcing the model to recognize these challenging adversarial styles, our model is gradually robust to the visual styles, thus boosting the generalization ability for novel target datasets.
Besides the typical CNN-based backbone, we also employ our StyleAdv method on large-scale pretrained vision transformer. Extensive experiments conducted on eight various target datasets show the effectiveness of our method.  Whether built upon ResNet or ViT, we achieve the new state of the art for CD-FSL.
Code is available at  \href{https://github.com/lovelyqian/StyleAdv-CDFSL}{https://github.com/lovelyqian/StyleAdv-CDFSL}.
\end{abstract}


\section{Introduction}
This paper studies the task of Cross-Domain Few-Shot Learning (CD-FSL) which addresses the Few-Shot Learning (FSL) problem across different domains.
As a general recipe for FSL, \textit{episode}-based meta-learning strategy has also been adopted for training CD-FSL models, e.g., FWT~\cite{tseng2020cross}, LRP~\cite{sun2020explanation}, ATA~\cite{wang2021cross}, and wave-SAN~\cite{fu2022wave}. Generally, to mimic the low-sample regime in testing stage, meta learning samples episodes for training the model. 
Each episode contains a small labeled \textit{support set} and an unlabeled \textit{query set}. Models learn meta knowledge by predicting the categories of images contained in the query set according to the support set. The learned meta knowledge generalizes the models to novel target classes directly.

Empirically, we find that the changes of visual appearances between source and target data is one of the key causes that leads to the domain gap in CD-FSL. Interestingly, wave-SAN~\cite{fu2022wave}, our former work,
shows that the domain gap issue can be alleviated by augmenting the visual styles of source images.
Particularly, wave-SAN proposes to augment the styles, in the form of Adaptive Instance Normalization (AdaIN)~\cite{huang2017arbitrary}, by randomly sampling two source episodes and exchanging their styles.
However, despite the efficacy of wave-SAN, such a na\"ive style generation method suffers from two limitations:
1) The swap operation makes the styles always be limited in the \textit{``real"} style set of the source dataset;
2) The limited real styles further lead to the generated styles too \textit{``easy''} to learn.
Therefore, a natural question is whether \textit{we can synthesize ``virtual'' and ``hard'' styles for learning a more robust CD-FSL model?}
Formally, we use ``real/virtual" to indicate whether the styles are originally presented in the set of source styles, and define ``easy/hard" as whether the new styles make meta tasks more difficult.

To that end, we draw inspiration from the adversarial training, and propose a novel meta \textbf{Style} \textbf{Adv}ersarial training method (\textbf{StyleAdv}) for CD-FSL. 
StyleAdv plays the minimax game in two iterative optimization loops of meta-training. 
Particularly, \textit{the inner loop} generates adversarial styles from the original source styles by adding perturbations. The synthesized adversarial styles are supposed to be more challenging for the current model to recognize, thus, increasing the loss.
Whilst \textit{the outer loop} optimizes the whole network by minimizing the losses of recognizing the images with both original and adversarial styles. Our ultimate goal is to enable learning a model that is robust to various styles, beyond the relatively limited and simple styles from the source data. This can potentially improve the generalization ability on novel target domains with visual appearance shifts.

Formally, we introduce a novel style adversarial attack method to support the inner loop of StyleAdv. 
Inspired yet different from the previous attack methods~\cite{goodfellow2014explaining, madry2017towards}, our style attack method perturbs and synthesizes the styles rather than image pixels or features. Technically,
we first extract the style from the input feature map, and include the extracted style in the forward computation chain to obtain its gradient for each training step.
After that, we synthesize the new style by adding a certain ratio of gradient to the original style.
Styles synthesized by our style adversarial attack method have the good properties of ``hard" and ``virtual".
Particularly, since we perturb styles in the opposite direction of the training gradients, 
our generation leads to the ``hard'' styles.
Our attack method results in totally ``virtual'' styles that are quite different from the original source styles.

Critically, our style attack method makes \textit{progressive style synthesizing}, with \textit{changing style perturbation ratios}, which makes it significantly different from vanilla adversarial attacking methods. 
Specifically, 
we propose a novel progressive style synthesizing strategy. 
The na\"ive solution of directly plugging-in perturbations is to
attack each block of the feature embedding module individually, which however, may results in large deviations of features from the high-level block. Thus, our strategy is to make the synthesizing signal of the current block be accumulated by adversarial styles from previous blocks.
On the other hand,
rather than attacking the models by fixing the attacking ratio, 
we synthesize new styles by randomly sampling the perturbation ratio from a candidate pool. 
This facilitates the diversity of the synthesized adversarial styles.
Experimental results have demonstrated the efficacy of our method: 1) our style adversarial attack method does synthesize more challenging styles, thus, pushing the limits of the source visual distribution; 2) our StyleAdv significantly improves the base model and outperforms all other CD-FSL competitors.

We highlight our StyleAdv is model-agnostic and complementary to other existing FSL or CD-FSL models, e.g., GNN~\cite{garcia2017few} and FWT~\cite{tseng2020cross}. 
More importantly, to benefit from the large-scale pretrained models, e.g., DINO~\cite{caron2021emerging}, we further explore adapting our StyleAdv to improve the Vision Transformer (ViT)~\cite{dosovitskiy2020image} backbone in a non-parametric way. 
Experimentally, we show that StyleAdv not only improves CNN-based FSL/CD-FSL methods, but also improves the large-scale pretrained ViT model.

Finally, we summarize our contributions.
1) A novel meta style adversarial training method, {termed} StyleAdv, is proposed for CD-FSL. By first perturbing the original styles and then forcing the model to learn from such adversarial styles, StyleAdv improves the robustness of CD-FSL models.
2) We present a novel style attack method with the novel progressive synthesizing strategy in changing attacking ratios. Diverse ``virtual'' and ``hard'' styles thus are generated.
3) Our method is complementary to existing FSL and CD-FSL methods; and we validate our idea on both CNN-based and ViT-based backbones.
4) Extensive results on eight unseen target datasets indicate that our StyleAdv outperforms previous CD-FSL methods, building a new SOTA result.

\section{Related Work}
\noindent\textbf{Cross-Domain Few-Shot Learning.} 
FSL which aims at freeing the model from reliance on massive labeled data has been studied for many years~\cite{snell2017prototypical,garcia2017few, ravi2016optimization, sun2019meta, tang2020blockmix, kumar2021protoda, xu2021learning, zhang2022progressive, tang2022learning}.
Particularly, some recent works, e.g., CLIP~\cite{radford2021learning}, CoOp~\cite{zhou2022learning}, CLIP-Adapter~\cite{gao2021clip}, Tip-Adapter~\cite{zhang2021tip}, and PMF~\cite{hu2022pushing} explore promoting the FSL with large-scale pretrained models.
Particularly, PMF contributes a simple pipeline and builds a SOTA for FSL.
As an extended task from FSL, CD-FSL~\cite{tseng2020cross, sun2020explanation, wang2021cross, fu2022wave, guo2020broader, li2021ranking, liang2021boosting, phoo2020self, islam2021dynamic, fu2021meta, cai2021damsl, guan2020large, zhengcross, fu2022generalized, fu2022me, zhuo2022tgdm} mainly solves the FSL across different domains.
Typical meta-learning based CD-FSL methods include FWT~\cite{tseng2020cross}, LRP~\cite{sun2020explanation}, ATA~\cite{wang2021cross}, AFA~\cite{hu2022adversarial}, and wave-SAN~\cite{fu2022wave}. 
Specifically, FWT and LRP tackle CD-FSL by refining batch normalization layers and using the explanation model to guide training.
ATA, AFA, and wave-SAN propose to augment the image pixels, features, and visual styles, respectively.
Several transfer-learning based CD-FSL methods, e.g., BSCD-FSL (also known as Fine-tune)~\cite{guo2020broader}, BSR~\cite{liu2020feature},
and NSAE~\cite{liang2021boosting}
have also been explored. These methods reveal that finetuning helps improving the performances on target datasets.
Other works that introduce extra data or require multiple domain datasets for training include STARTUP~\cite{phoo2020self}, Meta-FDMixup~\cite{fu2021meta},  Me-D2N~\cite{fu2022me}, TGDM~\cite{zhuo2022tgdm}, TriAE~\cite{guan2020large}, and DSL~\cite{hu2021switch}.

\noindent\textbf{Adversarial Attack.} The adversarial attack aims at misleading models by adding some bespoke perturbations to input data. 
To generate the perturbations effectively, lots of adversarial attack methods have been proposed~\cite{goodfellow2014explaining,madry2017towards,moosavi2016deepfool, duan2020adversarial,li2014feature,laidlaw2019functional,xie2020adversarial,xu2021towards}.
Most of the works~\cite{goodfellow2014explaining, madry2017towards, moosavi2016deepfool, duan2020adversarial} attack the image pixels. Specifically,  FGSM~\cite{goodfellow2014explaining} and PGD~\cite{madry2017towards} are two most classical and famous attack algorithms. Several works~\cite{li2014feature,laidlaw2019functional,zheng2021rectifying} attack the feature space.
Critically, few methods~\cite{xu2021towards} attack styles. 
Different from these works that aim to mislead the models, we perturb the styles to tackle the visual shift issue for CD-FSL.

\noindent\textbf{Adversarial Few-Shot Learning.}
Several attempts~\cite{li2020adversarial, wang2021fast, li2019defensive, goldblum2020adversarially, shen2019learning} that explore adversarial learning for FSL have been made. 
Among them, MDAT~\cite{li2019defensive}, AQ~\cite{goldblum2020adversarially}, and MetaAdv~\cite{wang2021fast} first attack the input image and then train the model using the attacked images to improve the defense ability against adversarial samples. Shen et al.~\cite{shen2019learning} attacks the feature of the episode to improve the generalization capability of FSL models.
Note that ATA~\cite{wang2021cross} and AFA~\cite{hu2022adversarial}, two CD-FSL methods, also adopt the adversarial learning. However, we are greatly different from them. ATA and AFA perturb image pixels or features, while we aim at bridging the visual gap by generating diverse hard styles.

\noindent\textbf{Style Augmentation for Domain Shift Problem.} Augmenting the style distribution for narrowing the domain shift issue has been explored in domain generation~\cite{zhou2021domain, li2022uncertainty, wang2021learning}, image segmentation~\cite{zhong2022adversarial, chen2022maxstyle}, person re-ID~\cite{zheng2019joint}, and CD-FSL~\cite{fu2022wave}. Concretely, MixStyle~\cite{zhou2021domain}, AdvStyle~\cite{zhong2022adversarial},
DSU~\cite{li2022uncertainty}, and wave-SAN~\cite{fu2022wave} synthesize styles without extra parameters via mixing, attacking, sampling from a Gaussian distribution, and swapping. 
MaxStyle~\cite{chen2022maxstyle} and L2D~\cite{wang2021learning} require additional network modules and complex auxiliary tasks to help generate the new styles.
Typically, AdvStyle~\cite{zhong2022adversarial} is the most related work to us. Thus, we highlight the key differences: 1) AdvStyle attacks styles on the image, while we attack styles on multiple feature spaces with a progressive attacking method; 2) AdvStyle uses the same task loss (segmentation) for attacking and optimization; in contrast, we use the classical classification loss to attack the styles, while utilize the task loss (FSL) to optimize the whole network.

\begin{figure*}
  \centering
  \includegraphics[width=1. \linewidth]{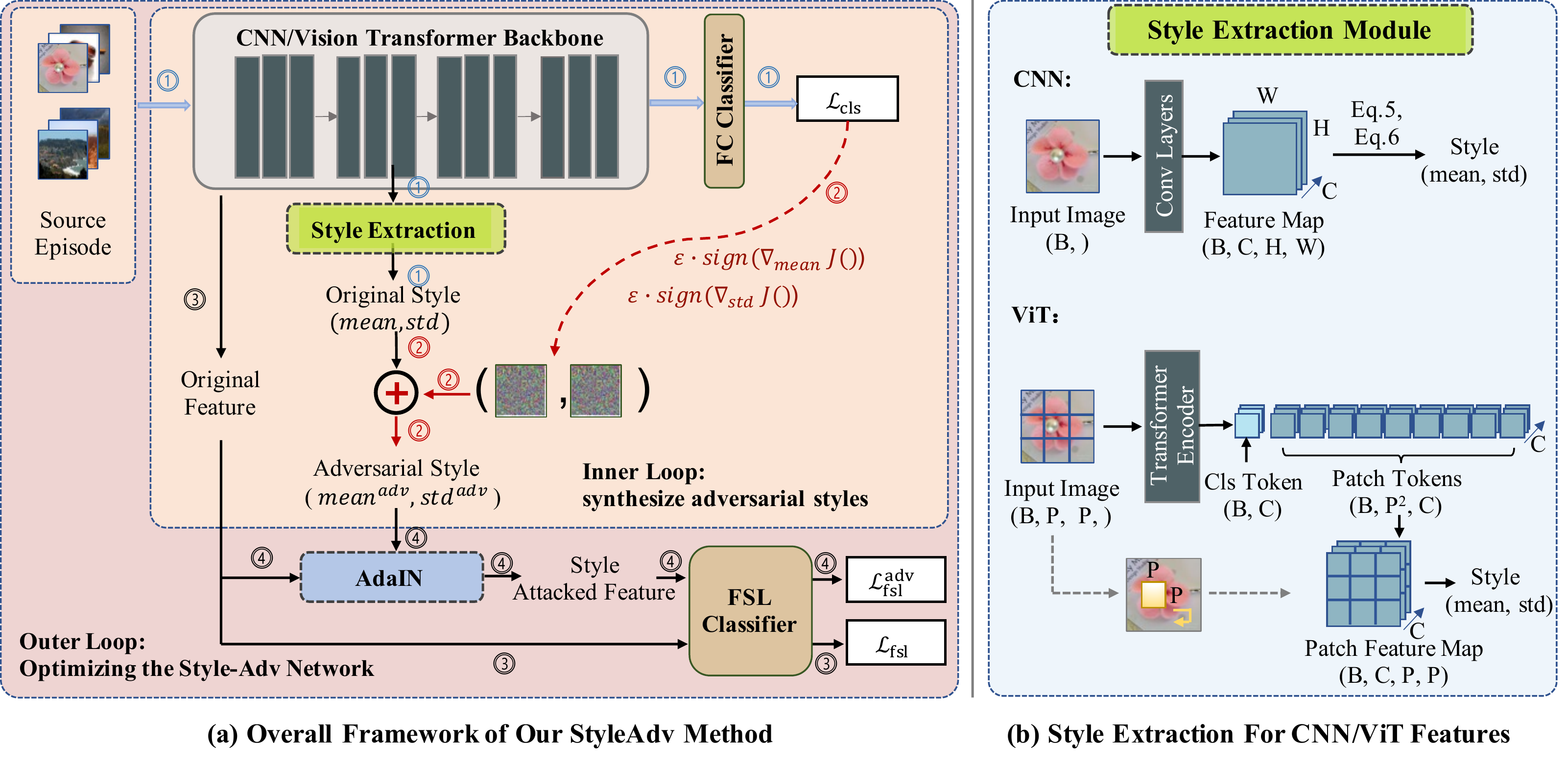}
  \vspace{-0.25in}
  \caption{\small (a): Overview of StyleAdv method. The inner loop synthesizes adversarial styles, while the outer loop optimizes the whole network.  (b): Style extraction for CNN-based and ViT-based features (illustration with B=1).
}
  \label{fig:framework}
  \vspace{-0.15in}
\end{figure*}

\section{StyleAdv: Meta Style Adversarial Training}

\noindent\textbf{Task Formulation.} 
Episode $\mathcal{T} = ((S, Q), Y)$ is randomly sampled as the input of each meta-task, where $Y$ represents the global class labels of the episode images with respect to $\mathcal{C}^{tr}$. Typically, each meta-task is formulated as an $N$-way $K$-shot problem. That is, for each episode $\mathcal{T}$, $N$ classes with $K$ labeled images are sampled as the support set $S$, and the same $N$ classes with another $M$ images are used to constitute the query set $Q$. The FSL or CD-FSL models predict the probability $P$ that the images in $Q$ belong to $N$ categories according to $S$. Formally, we have $|S| = NK$, $|Q| = NM$, $|P| = NM\times N$. 

\noindent\textbf{FGSM and PGD Attackers.}
We briefly summarize the algorithms for FGSM~\cite{goodfellow2014explaining} and PGD~\cite{madry2017towards}, two most famous attacking methods.
Given image $x$ with label $y$, 
FGSM attacks the $x$ by adding a ratio $\epsilon$ of signed gradients with respect to the $x$ resulting in the adversarial image $x^{adv}$ as,
\begin{equation}\label{eq:fgsm-std}
x^{adv} = x + \epsilon \cdot sign(\nabla_{x}J(\theta, x, y)),
\end{equation}
where $J(\cdot)$ and $\theta$ denote the object function and the learnable parameters of a classification model. PGD can be regarded as a variant of FGSM. Different from the FGSM that only attacks once, PGD attacks the image in an iterative way and sets a random start (abbreviated as RT) for $x$ as,
\begin{equation}\label{eq:pgd-std1}
x_{0}^{adv}  = x + k_{RT} \cdot \mathcal{N}(0,I),
\end{equation}
\begin{equation}\label{eq:pgd-std2}
x_{t}^{adv} = x_{t-1}^{adv} + \epsilon \cdot sign(\nabla_{x}J(\theta, x, y)),
\end{equation}
where $k_{RT}$, $\epsilon$ are hyper-parameters. $\mathcal{N}$ is Gaussian noises.

\subsection{Overview of  Meta Style Adversarial Learning}
To alleviate the performance degradation caused by the changing visual appearance, we tackle CD-FSL by promoting the robustness of models on recognizing various styles.
Thus, we expose our FSL model to some challenging virtual styles beyond the image styles that existed in the source dataset.
To that end, we present the novel StyleAdv adversarial training method. Critically, rather than adding perturbations to image pixels, we particularly focus on adversarially perturbing the styles.
The overall framework of our StyleAdv is illustrated in Figure~\ref{fig:framework}.
Our StyleAdv contains a CNN/ViT backbone $E$, a global FC classifier $f_{cls}$, and a FSL classifier $f_{fsl}$ with learnable parameters $\theta_{E},\theta_{cls}, \theta_{fsl}$, respectively. Besides, our core style attack method, a novel style extraction module, and the AdaIN are also included.

Overall, we learn the StyleAdv by solving a minimax game.  Specifically, the minimax game shall involve two iterative optimization loops in each meta-train step. 
Particularly,
\begin{itemize}[leftmargin=*]
\item
\textbf{Inner loop:} synthesizing new adversarial styles by attacking the original source styles; the generated styles will increase the loss of the current network.
\item 
\textbf{Outer loop:} optimizing the whole network by classifying source images with both original and adversarial styles; this process will decrease the loss.
\end{itemize}

\subsection{Style Extraction from CNNs and ViTs}\label{sec:style}
\noindent\textbf{Adaptive Instance Normalization (AdaIN).} 
We recap the vanilla AdaIN~\cite{huang2017arbitrary} proposed for CNN in style transfer. Particularly, AdaIN reveals that the instance-level mean and standard deviation (abbreviated as \textit{mean} and \textit{std}) convey the style information of the input image. Denoting the \textit{mean} and \textit{std} as $\mu$ and $\sigma$, AdaIN (denoted as $\mathcal{A}$) reveals that the style of $F$ can be transfered to that of $F_{tgt}$ by replacing the original style $(\mu, \sigma)$ with the target style $(\mu_{tgt}, \sigma_{tgt})$:

\begin{equation}\label{eq:adain}
\mathcal{A}(F,\mu_{tgt}, \sigma_{tgt})=\sigma_{tgt}\frac{F-\mu(F)}{\sigma(F)}+\mu_{tgt}.
\end{equation}

\noindent\textbf{Style Extraction for CNN Features.} 
As shown in the upper part of  Figure~\ref{fig:framework} (b), let $F\in \mathcal{R}^{B\times C \times H \times W}$ indicates the input feature batch, where $B$, $C$, $H$, and $W$ denote the batch size, channel, height, and width of the feature $F$, respectively. 
As in AdaIN, the \textit{mean} $\mu$ and \textit{std} $\sigma$ of $F$ are defined as:
	\begin{equation}\label{eq:mu}
	\mathrm{\mu(F)_{b,c}}=\frac{1}{HW}\sum_{h=1}^{H}\sum_{w=1}^{W}F_{b,c,h,w} ,
	\end{equation}
	\begin{equation}\label{eq:sigma}
	\negthickspace \negthickspace \mathrm{\sigma(F)_{b,c}}=\sqrt{\frac{1}{HW}\sum_{h=1}^{H}\sum_{w=1}^{W}(F_{b,c,h,w}-\mathrm{\mu_{b,c}}(F))^{2}+\epsilon} ,
	\end{equation}
 where $\mu, \sigma \in \mathcal{R}^{B\times C}$.
 
\noindent\textbf{Meta Information Extraction for ViT Features.}
We explore extracting the meta information of the ViT features
as the manner of CNN. 
Intuitively, such meta information can be regarded as a unique ``style" of ViTs.
As shown in Figure~\ref{fig:framework} (b), we take an input batch data with image split into $P\times P$ patches as an example. The ViT encoder will encode the batch patches into a class (cls) token ($F_{cls} \in \mathcal{R}^{B\times C}$) and a patch tokens ($F_{0} \in \mathcal{R}^{B\times P^{2} \times C}$). To be compatible with AdaIN, we reshape the $F_{0}$ as $F\in \mathcal{R}^{B\times C \times P \times P}$. At this point, we can calculate the meta information for patch tokens $F$ as in Eq.~\ref{eq:mu} and Eq.~\ref{eq:sigma}. 
Essentially, note that the transformer integrates the positional embedding into the patch representation, the spatial relations thus could be considered still hold in the patch tokens. This supports us to reform the patch tokens $F_{0}$ as a spatial feature map $F$. To some extent, this can also be achieved by applying the convolution on the input data via a kernel of size $P \times P$ (as indicated by dashed arrows in Figure~\ref{fig:framework} (b)).

\subsection{Inner Loop: Style Adversarial Attack Method}\label{sec:styleAttackMethod}
We propose a novel style adversarial attack method -- Fast Style Gradient Sign Method (Style-FGSM) to accomplish the inner loop. 
As shown in Figure~\ref{fig:framework}, given an input source episode $(\mathcal{T}, Y)$, we first forward it into the backbone $E$ and the FC classifier $f_{cls}$ producing the global classification loss $\mathcal{L}_{cls}$ (as illustrated in the $\normalsize{\textcircled{\scriptsize{1}}}\normalsize$ paths). 
During this process, a key step is to make the gradient of the style available. To achieve that, let $F_{\mathcal{T}}$ denotes the features of $\mathcal{T}$, we  obtain the style ($\mu$, $\sigma$) of $F_{\mathcal{T}}$ as in Sec.~\ref{sec:style}. After that, we reform the original episode feature as $\mathcal{A}(F_{\mathcal{T}}, \mu, \sigma$). And the reformed feature is actually used for the forward propagation. In this way, we include $\mu$ and $\sigma$ in our forward computation chain; and thus, we could access the gradients of them.

With the gradients in $\normalsize{\textcircled{\scriptsize{2}}}\normalsize$ paths, we then attack $\mu$ and $\sigma$ as FGSM does -- adding a small ratio $\epsilon$ of the signed gradients with respect to  $\mu$ and $\sigma$, respectively. 
\begin{equation}\label{eq:styleAttack-mean}\small
\begin{split}
\mu^{adv}  = 
\mu + \epsilon \cdot
 \mathrm{sign}(\nabla_{\mu}J(\theta_{E}, \theta_{f_{cls}}, 
\mathcal{A}
(F_{\mathcal{T}}, \mu, \sigma), Y)), 
\end{split}
\end{equation}
\begin{equation}\label{eq:styleAttack-std}
\sigma^{adv}  = \sigma +  \epsilon \cdot
\mathrm{sign}(\nabla_{\sigma}J(\theta_{E}, \theta_{f_{cls}}, \mathcal{A}(F_{\mathcal{T}}, \mu, \sigma), Y)), 
\end{equation}
where the $J()$ is the cross-entropy loss between  classification predictions and  ground truth $Y$, i.e., $\mathcal{L}_{cls}$.
Inspired by the random start of PGD, we also add random noises $k_{RT} \cdot \mathcal{N}(0,I)$ to $\mu$ and $\sigma$ before attacking.  $\mathcal{N}(0,I)$ refers to  Gaussian noises and $k_{RT}$ is a hyper-parameter.
Our Style-FGSM enables us to generate both ``virtual" and ``hard" styles.

\noindent\textbf{Progressive Style Synthesizing Strategy:}
To prevent the high-level adversarial feature from deviating, we propose to apply our style-FGSM in a progressive strategy. 
Concretely, the embedding module $E$ has three blocks $E_{1}$, $E_{2}$, and $E_{3}$,
with the corresponding features $F_{1}$, $F_{2}$, and $F_{3}$.
For the first block, we use ($\mu_{1}$, $\sigma_{1}$) to denote the original styles of $F_{1}$.
The adversarial styles $(\mu_{1}^{adv}, \sigma_{1}^{adv})$ are obtained directly as in Eq.~\ref{eq:styleAttack-mean} and Eq.~\ref{eq:styleAttack-std}. For subsequent blocks, the attack signals on the current block $i$ are those accumulated from the block $1$ to block $i-1$. 
Take the second block as an example, the block feature $F_{2}$ is not simply extracted by $E_{2}(F_{1})$. Instead, we have $F_{2}^{'} = E_{2}(F_{1}^{adv})$, where $F_{1}^{adv} = \mathcal{A}(F_{1}, \mu_{1}^{adv}, \sigma_{1}^{adv})$. Attacking on $F_{2}^{'}$ results in the adversarial styles $(\mu_{2}^{adv}, \sigma_{2}^{adv})$. Accordingly, we generate $(\mu_{3}^{adv}, \sigma_{3}^{adv})$ for the last block.
The illustration of the progressive attacking strategy is attached in the Appendix.

\noindent\textbf{Changing Style Perturbation Ratios: }
Different from the vanilla FGSM~\cite{goodfellow2014explaining} or PGD~\cite{madry2017towards}, our style attacking algorithm is expected to synthesize new styles with diversity. Thus, instead of using a fixed attacking ratio $\epsilon$, we randomly sample  $\epsilon$ from a candidate list $\epsilon_{list}$ as the current attacking ratio.
Despite the randomness of $\epsilon$, we still synthesize styles in a more challenging direction, $\epsilon$ only affects the extent.

\subsection{Outer Loop: Optimize the StyleAdv Network}
For each meta-train iteration with clean episode $\mathcal{T}$ as input, our inner loop produces adversarial styles $(\mu_{1}^{adv}, \sigma_{1}^{adv})$, $(\mu_{2}^{adv}, \sigma_{2}^{adv})$, and $(\mu_{3}^{adv}, \sigma_{3}^{adv})$.
As in Figure~\ref{fig:framework}, the goal of the outer loop is to optimize the whole StyleAdv with both the clean feature $F$ and the style attacked feature ${F}^{adv}$ utilized as the training data.
Typically, the clean episode feature $F$ can be obtained directly as $E(\mathcal{T})$ as in $\normalsize{\textcircled{\scriptsize{3}}}\normalsize$ paths.

In $\normalsize{\textcircled{\scriptsize{4}}}\normalsize$ paths, we obtain the $F^{adv}$ by transferring the original style of $F$ to the corresponding adversarial attacked styles.
Similar with the progressive style-FGSM, we have $F_{1}^{adv} = \mathcal{A}(E_{1}(\mathcal{T}) , \mu_{1}^{adv}, \sigma_{1}^{adv})$, $F_{2}^{adv} = \mathcal{A}(E_{2}(F_{1}^{adv}), \mu_{2}^{adv}, \sigma_{2}^{adv})$, and $F_{3}^{adv} = \mathcal{A}(E_{3}(F_{2}^{adv}), \mu_{3}^{adv}, \sigma_{3}^{adv})$. Finally, $F^{adv}$ is obtained by applying an average pooling layer to $F_{3}^{adv}$.
A skip probability $p_{skip}$ is set to decide whether to skip {the current attacking}. 
Conducting FSL tasks for both the clean feature $F$ and style attacked feature $F^{adv}$ results in two FSL predictions $P_{fsl}$, $P_{fsl}^{adv}$, and two FSL classification losses $\mathcal{L}_{fsl}$, $\mathcal{L}_{fsl}^{adv}$.

Further, despite the styles of $F^{adv}$ shifts from that of $F$, we encourage that the semantic content should be still consistent as in wave-SAN~\cite{fu2022wave}. Thus we add a consistent constraint to the predictions of $P_{fsl}$ and $P_{fsl}^{adv}$ resulting in the consistent loss $\mathcal{L}_{cons}$ as,
\begin{equation}
    \mathcal{L}_{cons} = \mathrm{KL}(P_{fsl}, P_{fsl}^{adv}),
\end{equation}
where $\mathrm{KL}()$ is  Kullback–Leibler divergence  loss. 
In addition, we have the global classification loss $\mathcal{L}_{cls}$.
This ensures that  $\theta_{cls}$  is optimized to provide correct gradients for style-FGSM. The final meta-objective of StyleAdv is as,
\begin{equation}\label{eq: train-loss}
\mathcal{L} = \mathcal{L}_{fsl} + \mathcal{L}_{fsl}^{adv} +\mathcal{L}_{cons} + \mathcal{L}_{cls}.
\end{equation}

Note that our StyleAdv is model-agnostic and orthogonal to existing FSL and CD-FSL methods.

\subsection{Network Inference}
\noindent\textbf{Applying StyleAdv Directly for Inference.}
Our StyleAdv facilitates making CD-FSL model more robust to style shifts. 
Once the model is meta-trained, we can employ it for inference directly by feeding the testing episode into the $E$ and the $f_{cls}$. The class with the highest probability will be taken as the predicted result.

\noindent\textbf{Finetuning StyleAdv Using Target Examples.}
As indicated in previous works~\cite{guo2020broader, liang2021boosting, liu2020feature, wang2021cross}, finetuning CD-FSL models on target examples helps improve the model performance. Thus, to further promote the performance of StyleAdv, we also equip it with the fintuning strategy forming an upgraded version (``StyleAdv-FT'').
Specifically, as in ATA-FT~\cite{wang2021cross}, for each novel testing episode, we augment the novel support set to form pseudo episodes as training data for tuning the meta-trained model.

\section{Experiments}\label{sec:experiments}
\noindent\textbf{Datasets.}
We take two CD-FSL benchmarks proposed in BSCD-FSL~\cite{guo2020broader} and FWT~\cite{tseng2020cross}. Both of them take mini-Imagenet~\cite{ravi2016optimization} as the source dataset. Two disjoint sets split from mini-Imagenet form $\mathcal{D}^{tr}$ and $\mathcal{D}^{eval}$. Totally eight datasets including ChestX~\cite{wang2017chestx}, ISIC~\cite{tschandl2018ham10000,codella2019skin}, EuroSAT~\cite{helber2019eurosat}, CropDisease~\cite{mohanty2016using}, CUB~\cite{wah2011caltech}, Cars~\cite{krause20133d}, Places~\cite{zhou2017places}, and Plantae~\cite{van2018inaturalist} are taken as novel target datasets. The former four datasets included in BSCD-FSL's benchmark cover medical images varying from X-ray to dermoscopic skin lesions, and natural images from satellite pictures to plant disease photos. While the latter four datasets that focus on more fine-grained concepts such as birds and cars are contained in FWT. These eight target datasets serve as testing set $\mathcal{D}^{te}$, respectively.

\noindent\textbf{Network Modules.}
For typical CNN based network, following previous CD-FSL methods~\cite{tseng2020cross, sun2020explanation, wang2021cross, fu2022wave}, ResNet-10~\cite{he2016deep} is selected as the embedding module while GNN~\cite{garcia2017few} is selected as the FSL classifier; For the emerging ViT based network, following PMF~\cite{hu2022pushing}, we use the ViT-small~\cite{dosovitskiy2020image} and the ProtoNet~\cite{snell2017prototypical} as the embedding module and the FSL classifier, respectively. Note that, the ViT-small is pretrained on ImageNet1K by DINO~\cite{caron2021emerging} as in PMF.  
The $f_{cls}$ is built by a fully connected layer.

\noindent\textbf{Implementation Details.}
The 5-way 1-shot and 5-way 5-shot settings are conducted. 
Taking ResNet10 as backbone, we meta train the network for 200 epochs, each epoch contains 120 meta tasks. Adam with a learning rate of $0.001$ is utilized as the optimizer.
Taking ViT-small as backbone, the meta train stage takes 20 epoch, each epoch contains 2000 meta tasks. The SGD with a initial learning rate of 5e-5 and 0.001 are used for optimize the $E()$ and the $f_{cls}$, respectively.
The $\epsilon_{list}$,  $k_{RT}$ of Style-FGSM attacker are set as $[0.8, 0.08, 0.008]$, $\frac{16}{255}$. The probability $p_{skip}$ of random skipping the attacking is chosen from $\left\{0.2,0.4\right\}$.
We evaluate our network with 1000 randomly sampled episodes and report average accuracy (\%) with a 95\% confidence interval.
Both the results of our ``StyleAdv'' and ``StyleAdv-FT'' are reported. The details of the finetuning are attached in Appendix.
ResNet-10 based models are trained and tested on a single GeForce GTX 1080, while ViT-small based models require a single NVIDIA GeForce RTX 3090.

\setlength{\tabcolsep}{3pt} 
\begin{table*}
   \scriptsize
    \centering 
    \begin{adjustbox}{width=\textwidth}
    \begin{tabular}{l|c| c | c |  cccc|cccc|c} %
    \toprule
       \textbf{1-shot} & \textbf{Backbone} & 
       \textbf{FT} & 
       \textbf{LargeP}  & 
       \texttt{\bf ChestX} & \texttt{\bf ISIC}& \texttt{\bf EuroSAT} & \texttt{\bf CropDisease} & \texttt{\bf CUB} &
       \texttt{\bf Cars}& \texttt{\bf Places} & \texttt{\bf Plantae} & {\textbf{Average}}\\
       \midrule
       GNN~\cite{garcia2017few}
       & RN10 & - & - & 22.00\mypm0.46  & 32.02\mypm0.66 & 63.69\mypm1.03 & 64.48\mypm1.08 & 45.69\mypm0.68   & 31.79\mypm0.51   & 53.10\mypm0.80   & 35.60\mypm0.56 & \cellcolor{blue!15}43.55\\ 
       
       FWT~\cite{tseng2020cross} & RN10 & - & -&  
       22.04\mypm0.44 & 31.58\mypm0.67 & 62.36\mypm1.05 & 66.36\mypm1.04 & 47.47\mypm0.75   & 31.61\mypm0.53   & 55.77\mypm0.79   & 35.95\mypm0.58  & \cellcolor{blue!15}44.14\\ 
       
       LRP~\cite{sun2020explanation}& RN10  & - & - &
       22.11\mypm0.20 &  30.94\mypm0.30 & 54.99\mypm0.50 & 59.23\mypm0.50 &  48.29\mypm0.51   & 32.78\mypm0.39   & 54.83 \mypm0.56    & 37.49 \mypm0.43 & \cellcolor{blue!15}42.58\\
       
       ATA~\cite{wang2021cross} & RN10 & - & - &
       22.10\mypm0.20 & 33.21\mypm0.40 & 61.35\mypm0.50 & 67.47\mypm0.50 & 45.00\mypm0.50   & 33.61\mypm0.40 & 53.57\mypm0.50   & 34.42\mypm0.40 & \cellcolor{blue!15}43.84\\
       
       AFA~\cite{hu2022adversarial} & RN10 & - & - & 22.92\mypm0.20 & 33.21\mypm0.30 & 63.12\mypm0.50 & 67.61\mypm0.50 & 46.86\mypm0.50 & 34.25\mypm0.40 & 54.04\mypm0.60 & 36.76\mypm0.40 & \cellcolor{blue!15}44.85\\

       wave-SAN~\cite{fu2022wave} & RN10 & - & - & \textbf{22.93\mypm0.49} & 33.35\mypm0.71 & 69.64\mypm1.09 & 70.80\mypm1.06 & \textbf{50.25\mypm0.74} & 33.55\mypm0.61 & 57.75\mypm0.82 & 40.71\mypm0.66 & \cellcolor{blue!15}47.37 \\

    \textbf{StyleAdv (ours)} & RN10 & - & - &
    \cellcolor{blue!15}22.64\mypm0.35 &
	\cellcolor{blue!15}\textbf{33.96\mypm0.57} &\cellcolor{blue!15}\textbf{70.94\mypm0.82} &
	\cellcolor{blue!15}\textbf{74.13\mypm0.78} & 
	\cellcolor{blue!15}48.49\mypm0.72 &
	\cellcolor{blue!15}\textbf{34.64\mypm0.57} &
	\cellcolor{blue!15}\textbf{58.58\mypm0.83} &
	\cellcolor{blue!15}\textbf{41.13\mypm0.67} & \cellcolor{blue!15}\textbf{48.06} \\
    
    \midrule
    
    ATA-FT~\cite{wang2021cross} & RN10 & Y & - & 22.15\mypm0.20 & 34.94\mypm0.40  & 68.62\mypm0.50  & 75.41\mypm0.50 & 46.23\mypm0.50 & \textbf{37.15\mypm0.40} & 54.18\mypm0.50 & 37.38\mypm0.40 & \cellcolor{blue!15} 47.01 \\
      
    \textbf{StyleAdv-FT (ours)} & RN10 & Y & - & 
    \cellcolor{blue!15}\textbf{22.64\mypm0.35} &\cellcolor{blue!15}\textbf{35.76\mypm0.52} &\cellcolor{blue!15}\textbf{72.92\mypm0.75} &\cellcolor{blue!15}\textbf{80.69\mypm0.28} &\cellcolor{blue!15}\textbf{48.49\mypm0.72} &\cellcolor{blue!15}35.09\mypm0.55 &\cellcolor{blue!15}\textbf{58.58\mypm0.83} &\cellcolor{blue!15}\textbf{41.13\mypm0.67} &\cellcolor{blue!15}\textbf{49.41}
    \\

    \midrule
    PMF$^\ast$~\cite{hu2022pushing} & ViT-small & Y & DINO/IN1K & 
    21.73\mypm0.30 &	30.36\mypm0.36 &	70.74\mypm0.63 &	80.79\mypm0.62 &	78.13\mypm0.66 &	37.24\mypm0.57 &	71.11\mypm0.71 &	53.60\mypm0.66 &  \cellcolor{blue!15}55.46
    \\
    
    \textbf{StyleAdv (ours)} & ViT-small & - & DINO/IN1K &  \cellcolor{blue!15}\textbf{22.92\mypm0.32} &\cellcolor{blue!15}\textbf{33.05\mypm0.44} &\cellcolor{blue!15}\textbf{72.15\mypm0.65} &\cellcolor{blue!15}\textbf{81.22\mypm0.61} &\cellcolor{blue!15}\textbf{84.01\mypm0.58} &\cellcolor{blue!15}\textbf{40.48\mypm0.57} &\cellcolor{blue!15}\textbf{72.64\mypm0.67} &\cellcolor{blue!15}\textbf{55.52\mypm0.66} &\cellcolor{blue!15}\textbf{57.75}\\ 
   
	\textbf{StyleAdv-FT (ours)} & ViT-small & Y & DINO/IN1K &\cellcolor{blue!15}\textbf{22.92\mypm0.32} &\cellcolor{blue!15}\textbf{33.99\mypm0.46} &\cellcolor{blue!15}\textbf{74.93\mypm0.58} &\cellcolor{blue!15}\textbf{84.11\mypm0.57} &\cellcolor{blue!15}\textbf{84.01\mypm0.58} &\cellcolor{blue!15}\textbf{40.48\mypm0.57} &\cellcolor{blue!15}\textbf{72.64\mypm0.67} &\cellcolor{blue!15}\textbf{55.52\mypm0.66} &\cellcolor{blue!15}\textbf{58.57}\\ 
	
	 \midrule
	 \midrule

	 \textbf{5-shot} & \textbf{Backbone} & \textbf{FT} & \textbf{LargeP}  & 
     \texttt{\bf ChestX} & \texttt{\bf ISIC}& \texttt{\bf EuroSAT} & \texttt{\bf CropDisease} & \texttt{\bf CUB} & \texttt{\bf Cars}& \texttt{\bf Places} & \texttt{\bf Plantae} & {\textbf{Average}}\\
     \midrule
     GNN~\cite{garcia2017few} 
     & RN10 & - & -& 
     25.27\mypm0.46 &  43.94\mypm0.67 & 83.64\mypm0.77 & 87.96\mypm0.67 & 62.25\mypm0.65   & 44.28\mypm0.63   & 70.84\mypm0.65   & 52.53\mypm0.59 & \cellcolor{blue!15}58.84 \\ 
     
     FWT~\cite{tseng2020cross}
     & RN10 & - & - & 
     25.18\mypm0.45  & 43.17\mypm0.70 & 83.01\mypm0.79 & 87.11\mypm0.67 & 66.98\mypm0.68 & 44.90\mypm0.64 & 73.94\mypm0.67 & 53.85\mypm0.62 & \cellcolor{blue!15}59.77\\
     
     LRP~\cite{sun2020explanation} & RN10 & - & -  & 
     24.53\mypm0.30  & 44.14\mypm0.40 & 77.14\mypm0.40  & 86.15\mypm0.40 & 64.44\mypm0.48 & 46.20\mypm0.46  & 74.45\mypm0.47 & 54.46\mypm0.46 & \cellcolor{blue!15}58.94\\
     
     ATA~\cite{wang2021cross} & 
     RN10 & - & - & 
     24.32\mypm0.40 & 44.91\mypm0.40 & 83.75\mypm0.40 & 90.59\mypm0.30 & 66.22\mypm0.50   & 49.14\mypm0.40 & 75.48\mypm0.40   & 52.69\mypm0.40 & \cellcolor{blue!15}60.89\\
     
     AFA~\cite{hu2022adversarial} & RN10 & - & - &
     25.02\mypm0.20 & \textbf{46.01\mypm0.40} & 85.58\mypm0.40 & 88.06\mypm0.30 & 68.25\mypm0.50 & 49.28\mypm0.50 & 76.21\mypm0.50 & 54.26\mypm0.40 & \cellcolor{blue!15}61.58\\

     wave-SAN~\cite{fu2022wave} &
     RN10 & - & - & 
     25.63\mypm0.49 & 44.93\mypm0.67 & 85.22\mypm0.71 & 89.70\mypm0.64 & \textbf{70.31\mypm0.67} & 46.11\mypm0.66 & 76.88\mypm0.63 & 57.72\mypm0.64 &\cellcolor{blue!15}62.06 \\
    
    \textbf{StyleAdv (ours)} & 
     RN10  & - & - &
     \cellcolor{blue!15}\textbf{26.07\mypm 0.37} &\cellcolor{blue!15}45.77\mypm 0.51 &
	\cellcolor{blue!15}\textbf{86.58\mypm 0.54} &
	\cellcolor{blue!15}\textbf{93.65\mypm 0.39} & 
	\cellcolor{blue!15}68.72\mypm 0.67 &
	\cellcolor{blue!15}\textbf{50.13\mypm 0.68} &
	\cellcolor{blue!15}\textbf{77.73\mypm 0.62} &
	\cellcolor{blue!15}\textbf{61.52\mypm 0.68} &  \cellcolor{blue!15}\textbf{63.77} \\

     \midrule 
     Fine-tune~\cite{guo2020broader} & RN10 & Y & - &  
     25.97\mypm0.41 & 48.11\mypm0.64 & 79.08\mypm0.61 & 89.25\mypm0.51 & 64.14\mypm0.77 & 52.08\mypm0.74 & 70.06\mypm0.74 & 59.27\mypm0.70
     & \cellcolor{blue!15}61.00\\
     
     ATA-FT~\cite{wang2021cross} & RN10 & Y & - & 
     25.08\mypm0.20 & 49.79\mypm0.40 & 89.64\mypm0.30 & 95.44\mypm0.20 & 69.83\mypm0.50 & 54.28\mypm0.50 & 76.64\mypm0.40 & 58.08\mypm0.40 & \cellcolor{blue!15}64.85 \\
     
     NSAE~\cite{liang2021boosting} & RN10 & Y & - &\textbf{27.10\mypm0.44} & 54.05\mypm0.63 & 83.96\mypm0.57 & 93.14\mypm0.47 & 68.51\mypm0.76 & 54.91\mypm0.74 & 71.02\mypm0.72 & 59.55\mypm0.74 & \cellcolor{blue!15}64.03 \\
     
     BSR~\cite{liu2020feature} & RN10 & Y & - &
     26.84\mypm0.44 & \textbf{54.42\mypm0.66} & 80.89\mypm0.61 & 92.17\mypm0.45 & 69.38\mypm0.76 & \textbf{57.49\mypm0.72} & 71.09\mypm0.68 & 61.07\mypm0.76 & \cellcolor{blue!15}64.17\\
     
    \textbf{StyleAdv-FT (ours)} & RN10 & Y & - & 
    \cellcolor{blue!15}26.24\mypm0.35&	\cellcolor{blue!15}53.05\mypm0.54&	\cellcolor{blue!15}\textbf{91.64\mypm0.43}&	\cellcolor{blue!15}\textbf{96.51\mypm0.28}&	\cellcolor{blue!15}\textbf{70.90\mypm0.63}&	\cellcolor{blue!15}56.44\mypm0.68&	\cellcolor{blue!15}\textbf{79.35\mypm0.61}&	\cellcolor{blue!15}\textbf{64.10\mypm0.64} & \cellcolor{blue!15}\textbf{67.28}\\
    
   \midrule
    PMF~\cite{hu2022pushing} & ViT-small & Y & DINO/IN1K & 
    \textbf{27.27} 
    & 50.12  
    & 85.98  
    & 92.96 & - & - & - & - & \cellcolor{blue!15} - \\
    \textbf{StyleAdv (ours)} & ViT-small & - & DINO/IN1K & \cellcolor{blue!15}26.97\mypm0.33 & 	\cellcolor{blue!15}47.73\mypm0.44 &	\cellcolor{blue!15}\textbf{88.57\mypm0.34} & 	\cellcolor{blue!15}\textbf{94.85\mypm0.31} &	\cellcolor{blue!15}\textbf{95.82\mypm0.27} &	\cellcolor{blue!15}\textbf{61.73\mypm0.62} &	\cellcolor{blue!15}\textbf{88.33\mypm0.40} &	\cellcolor{blue!15}\textbf{75.55\mypm0.54} & \cellcolor{blue!15}\textbf{72.44}\\ 
    
	\textbf{StyleAdv-FT (ours)} & ViT-small & Y & DINO/IN1K & \cellcolor{blue!15}26.97\mypm0.33& \cellcolor{blue!15}\textbf{51.23\mypm0.51}&	\cellcolor{blue!15}\textbf{90.12\mypm0.33}&\cellcolor{blue!15}\textbf{95.99\mypm0.27}&\cellcolor{blue!15}\textbf{95.82\mypm0.27} &\cellcolor{blue!15}\textbf{66.02\mypm0.64} &\cellcolor{blue!15}\textbf{88.33\mypm0.40} &\cellcolor{blue!15}\textbf{78.01\mypm0.54} & \cellcolor{blue!15}\textbf{74.06}
	\\ 
    
	\bottomrule   
 	\end{tabular}
    \end{adjustbox}
    \vspace{-0.1in}
    \caption{
    \small
    Results of 5-way 1-shot/5-shot tasks.  ``FT'' means whether the finetuning stage is employed. ``LargeP"' represents if large pretrained models are used for model initialization. ``RN10" is short for ``ResNet-10''. $\ast$ denotes results are reported by us.  Results perform best are bolded. 
    Whether based on ResNet-10 or ViT-small, our method outperforms other competitors significantly.
    \label{tab:main-result-new}}
    \vspace{-0.2in}
\end{table*}

\subsection{Comparison with the SOTAs}
We compare our StyleAdv/StyleAdv-FT against several most representative and competitive CD-FSL methods. 
Concretly, with the ResNet-10 (abbreviated as RN10) as backbone, totally nine methods including GNN~\cite{garcia2017few}, FWT~\cite{tseng2020cross}, LRP~\cite{sun2020explanation}, ATA~\cite{wang2021cross}, AFA~\cite{hu2022adversarial}, wave-SAN~\cite{fu2022wave}, Fine-tune~\cite{guo2020broader}, NSAE~\cite{liang2021boosting}, and BSR~\cite{liu2020feature} are introduced as our competitors. Among them, the former six competitors are meta-learning based method that used for inference directly, thus we compare our ``StyleAdv'' against them for a fair comparison.
Typically, the GNN~\cite{garcia2017few} works as a base model. 
The Fine-tune~\cite{guo2020broader}, NSAE~\cite{liang2021boosting}, BSR~\cite{liu2020feature}, and ATA-FT~\cite{wang2021cross} (formed by finetuning ATA) all require finetuning model during inference, thus our ``StyleAdv-FT'' is used.
With the ViT as backbone, the most recent and competitive PMF (SOTA method for FSL) is compared. For fair comparisons, we follow the same pipeline proposed in PMF~\cite{hu2022pushing}.
Note that we promote CD-FSL models with only one single source domain. Those methods that use extra training datasets, e.g., STARTUP~\cite{phoo2020self}, meta-FDMixup~\cite{fu2021meta}, and DSL~\cite{hu2021switch} are not considered.
The comparison results are given in Table~\ref{tab:main-result-new}.

For all results, our method outperforms all the listed CD-FSL competitors significantly and builds a new state of the art.
Our StyleAdv-FT (ViT-small) on average achieves $58.57\%$ and $74.06\%$ on 5-way 1-shot and 5-shot, respectively.
Our StyleAdv (RN10) and StyleAdv-FT (RN10) also beats all the meta-learning based or transfer-learning (finetuning) based methods.
Besides of the state-of-the-art accuracy, we also have other worth-mentioning observations.
1) We show that our StyleAdv method is a general solution for both CNN-based models and ViT-based models. 
Typically, based on ResNet10, our StyleAdv and StyleAdv-FT improve the base GNN by up to $4.93\%$ and $8.44\%$ on 5-shot setting. Based on ViT-small, at most cases, our StyleAdv-FT outperforms the PMF by a clear margin.
More results of building StyleAdv upon other FSL or CD-FSL methods can be found in the Appendix.
2) Comparing FWT, LRP, ATA, AFA, waveSAN, and our StyleAdv, we find that StyleAdv performs best, followed by wave-SAN, then comes the AFA, ATA, FWT, and LRP. This phenomenon indicates that tackling CD-FSL by solving the visual shift problem is indeed more effective than other perspectives, e.g., adversarial training by perturbing the image features (AFA) or image pixels (ATA), transforms the normalization layers in FWT, and explanation guided training in LRP.
3) For the comparison between StyleAdv and wave-SAN that both tackles the visual styles, we notice that StyleAdv outperforms the wave-SAN in most cases. 
This demonstrates that the styles generated by our StyleAdv are more conducive to learning robust CD-FSL models than the style augmentation method proposed in wave-SAN.
This justifies our idea of synthesizing more challenging (``hard and virtual'') styles. 
4) Overall, the large-scale pretrained model promotes the CD-FSL obviously. Take 1-shot as an example, StyleAdv-FT (ViT-small) boosts the StyleAdv-FT (RN10) by $9.16\%$ on average. However, we show that the performance improvement varies greatly on different target domains. Generally, for target datasets with relative small domain gap, e.g., CUB and Plantae, models benefit a lot; otherwise, the improvement is limited.
5) We also find that under the cross-domain scenarios, finetuning model on target domain, e.g., NSAE, BSR do show an advantange over purely meta-learning based methods, e.g., FWT, LRP, and wave-SAN. However, to finetune model using extremely few examples, e.g., 5-way 1-shot is much harder than on relatively larger shots. This may explain why those finetune-based methods do not conduct experiments on 1-shot setting.

\begin{figure}[ht]
  \centering
  \includegraphics[width=0.85 \linewidth]{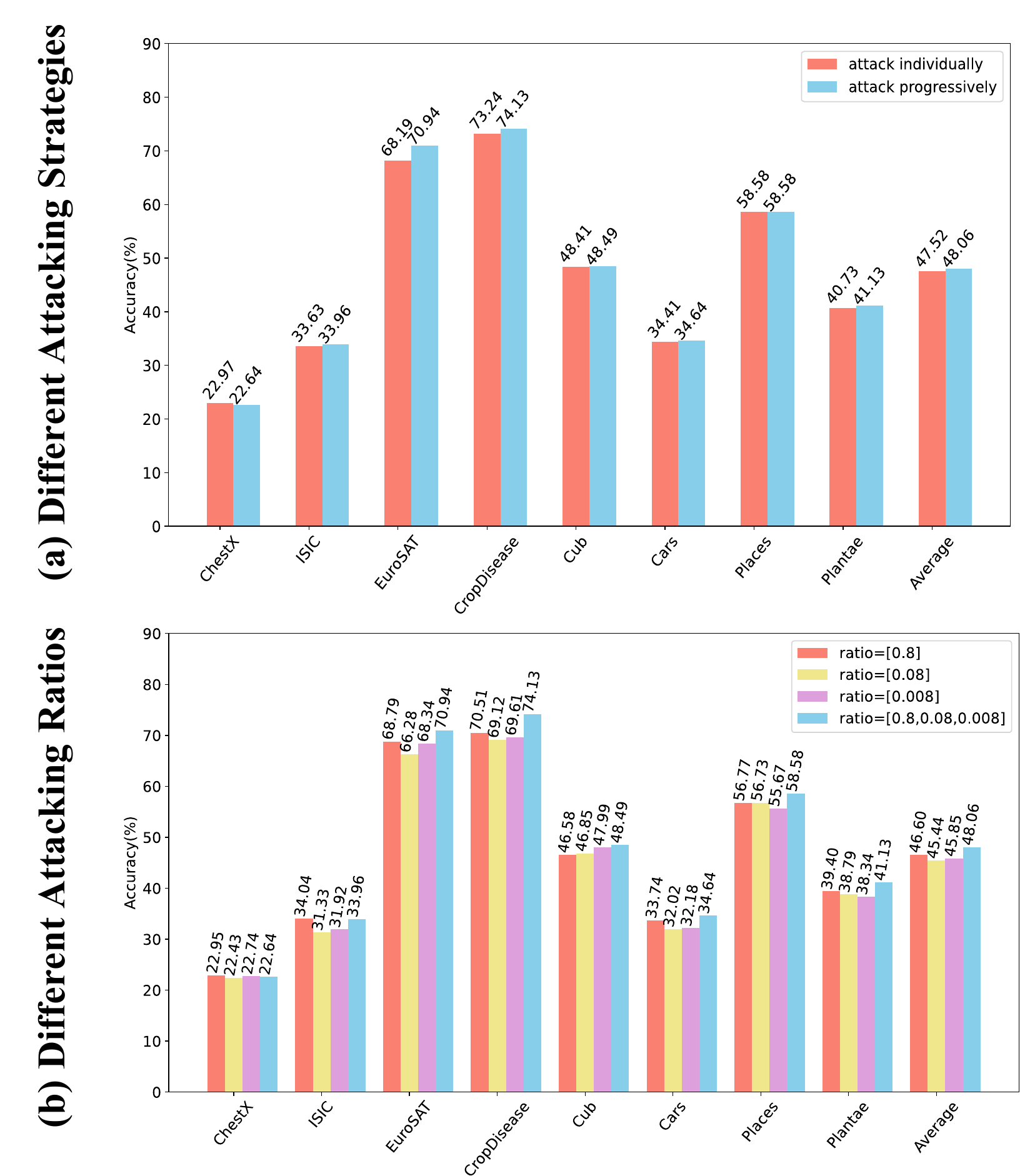}
  \vspace{-0.15in}
  \caption{ \small
  Effectiveness of the progressive style synthesizing strategy and the changing style perturbation ratios. The 5-way 1-shot results are reported. Models are built on ResNet10 and GNN.
  }
  \label{fig:abla}
  \vspace{-0.25in}
\end{figure}

\noindent \textbf{Effectiveness of Style-FGSM Attacker.}
To show the advantages of our progressive style synthesizing strategy and attacking with changing perturbation ratios, we compare our Style-FGSM against several variants and report the results in Figure~\ref{fig:abla}.
Specifically, for Figure~\ref{fig:abla} (a), we compare our style-FGSM against the variant that attacks the blocks individually. Results show that attacking in a progressive way exceeds the naive individual strategy in most cases.
For Figure~\ref{fig:abla} (b), to demonstrate how the performance will be affected by fixed attacking ratios, we also conduct experiments with different $\epsilon_{list}$. Since we set the $\epsilon_{list}$ as $[0.8, 0.08, 0.008]$, three different choices including $[0.8]$, $[0.08]$, and $[0.008]$ are selected. From the results, we first notice that the best result can be reached by a single fixed ratio. However, sampling the attacking ratio from a pool of candidates achieves the best result in most cases.

\subsection{More Analysis}
\begin{figure}
  \centering
  \includegraphics[width=0.9 \linewidth]{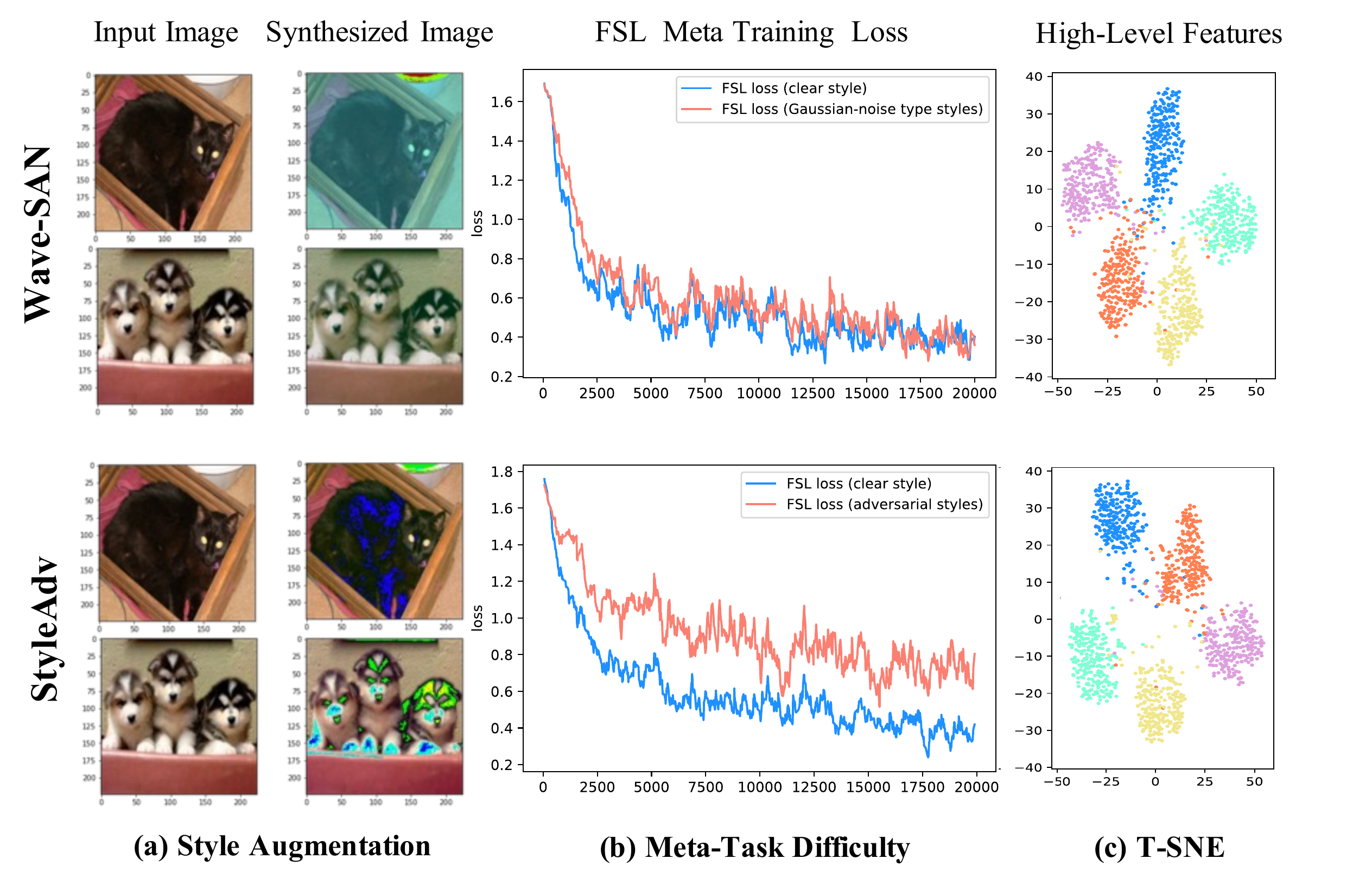}
  \vspace{-0.1in}
  \caption{\small
  \textbf{Visualization of wave-SAN and StyleAdv.}
  (a): synthesized images; (b): meta-training losses; (c): T-SNE results.
  }
  \label{fig:vis-new}
  \vspace{-0.3in}
\end{figure}

\noindent\textbf{Visualization of Hard Style Generation.}
To help understand the ``hard" style generation of our method intuitively, as in Figure~\ref{fig:vis-new}, we make several visualizations comparing StyleAdv against the wave-SAN.
1) As in Figure~\ref{fig:vis-new} (a), we show the stylized images generated by wave-SAN and our StyleAdv. 
The visualization is achieved by applying the style augmentation methods to input images. Specifically, for wave-SAN, the style is swapped with another randomly sampled source image; 
for StyleAdv, the results of attacking style with $\epsilon = 0.08$ are given.
We observe that wave-SAN tends to exchange the global visual appearance, e.g., the color of the input image randomly. By contrast, StyleAdv prefers to  disturb the important regions  that are key to recognizing the image category.  
For example, the fur of the cat and the key parts (face and feet) of the dogs.
These observations intuitively support our claim that our StyleAdv synthesize more harder styles than wave-SAN.
2) To quantitatively evaluate whether our StyleAdv introduces more challenging styles into the training stage, as in Figure~\ref{fig:vis-new} (b), we visualize the meta-training loss. 
Results reveal that the perturbed losses of wave-SAN oscillate around the original loss, while StyleAdv increases the original loss obviously. These phenomenons further validate that we perturb data towards a more difficult direction thus pushing the limits of style generation to a large extent.
3) To further show the advantages of StyleAdv over wave-SAN, as shown in Figure~\ref{fig:vis-new} (c), we visualize the high-level features extracted by the meta-trained wave-SAN and StyleAdv. Five classes (denoted by different colors) of mini-Imagenet are selected. T-SNE is used for reducing the feature dimensions. 
Results demonstrate that StyleAdv enlarges the inter-class distances making classes more distinguishable. 

\setlength{\tabcolsep}{3pt}
\begin{table*}
   \scriptsize
    \centering  
    \begin{adjustbox}{width=\textwidth}
    \begin{tabular}{l|c|cccc|cccc|c} %
    \toprule
	  & \textbf{Attack Target} & 
     \texttt{\bf ChestX} & \texttt{\bf ISIC}& \texttt{\bf EuroSAT} & \texttt{\bf CropDisease} & \texttt{\bf CUB} & \texttt{\bf Cars}& \texttt{\bf Places} & \texttt{\bf Plantae} & {\textbf{Average}}\\
     \midrule
     1-shot
     &  Image & 
    \textbf{22.71\mypm0.35} & 
	33.00\mypm0.53 & 
	67.00\mypm0.82 & 
	72.65\mypm0.75 & 
	48.15\mypm0.72 & 
	34.40\mypm0.60 & 
	57.89\mypm0.83 & 	
	39.85\mypm0.64 &  \cellcolor{blue!15}46.96
\\ 

     &  Feature & 22.55\mypm0.35 & 
	32.95\mypm0.53 & 
	68.71\mypm0.81 & 
	70.86\mypm0.78 & 
	46.52\mypm0.70 & 
	34.07\mypm0.54 & 
	56.68\mypm0.81 & 
	39.62\mypm0.62 & \cellcolor{blue!15}46.50
\\ 
     &  \textbf{Style (ours)} &  
    \cellcolor{blue!15}22.64\mypm0.35 &
	\cellcolor{blue!15}\textbf{33.96\mypm0.57} &
	\cellcolor{blue!15}\textbf{70.94\mypm0.82} &
	\cellcolor{blue!15}\textbf{74.13\mypm0.78} &
	\cellcolor{blue!15}\textbf{48.49\mypm0.72} &
	\cellcolor{blue!15}\textbf{34.64\mypm0.57} &
	\cellcolor{blue!15}\textbf{58.58\mypm0.83} &
	\cellcolor{blue!15}\textbf{41.13\mypm0.67} & \cellcolor{blue!15}\textbf{48.06}\\ 
     \midrule
     
    5-shot
     &  Image & 24.92\mypm0.36 & 	
     42.63\mypm0.47 & 	
     84.18\mypm0.54 & 
	90.31\mypm0.47 & 
	66.37\mypm0.65 & 
	47.46\mypm0.67 & 
	75.94\mypm0.62 & 
	57.33\mypm0.65 & \cellcolor{blue!15}61.14
\\ 

     &  Feature & 
    25.55\mypm0.37 & 
	43.71\mypm0.50 & 	
	84.22\mypm0.55 & 
	91.71\mypm0.44 & 
	67.31\mypm0.67 & 
	\textbf{50.26\mypm0.67} & 
	76.46\mypm0.65 & 
	57.39\mypm0.63 & \cellcolor{blue!15}62.08\\ 
	
     &  \textbf{Style (ours)}  &  \cellcolor{blue!15}
     \textbf{26.07\mypm 0.37} &\cellcolor{blue!15}\textbf{45.77\mypm 0.51} &
	\cellcolor{blue!15}\textbf{86.58\mypm 0.54} &
	\cellcolor{blue!15}\textbf{93.65\mypm 0.39} & 
	\cellcolor{blue!15}\textbf{68.72\mypm 0.67} &
	\cellcolor{blue!15}50.13\mypm 0.68 &
	\cellcolor{blue!15}\textbf{77.73\mypm 0.62} &
	\cellcolor{blue!15}\textbf{61.52\mypm 0.68} & \cellcolor{blue!15}\textbf{63.77}\\ 
	 \bottomrule   
 	\end{tabular}
    \end{adjustbox}
    \vspace{-0.15in}
    \caption{\small
    Comparison results (\%) of attacking image, feature, and styles.  
    Models build upon ResNet10 and GNN classifier.}
    \label{tab:AttackObject}
     \vspace{-0.1in}
\end{table*}

\setlength{\tabcolsep}{3pt}
\begin{table*}
   \scriptsize
    \centering  
    \begin{adjustbox}{width=\textwidth}
    \begin{tabular}{l|c|cccc|cccc|c} %
    \toprule
	 & \textbf{Augment Method} & 
     \texttt{\bf ChestX} & \texttt{\bf ISIC}& \texttt{\bf EuroSAT} & \texttt{\bf CropDisease} & \texttt{\bf CUB} & \texttt{\bf Cars}& \texttt{\bf Places} & \texttt{\bf Plantae} & {\textbf{Average}}\\
     \midrule
     1-shot 
     & StyleGaus$\dagger$ & 
     22.37\mypm0.35&
	31.48\mypm0.52&
	65.71\mypm0.82&
	69.25\mypm0.80&
	46.32\mypm0.72&
	32.69\mypm0.54&
	55.48\mypm0.79&
	37.27\mypm0.61& \cellcolor{blue!15}45.07
\\
     
     & MixStyle~\cite{zhou2021domain} & 
     22.43\mypm0.35&	
     33.21\mypm0.53&	
     67.35\mypm0.80&	
     68.80\mypm0.82&		
     47.08\mypm0.73&	
     33.39\mypm0.58&	
     56.12\mypm0.78&	
     38.03\mypm0.62& \cellcolor{blue!15}45.80 \\ 
     
     & AdvStyle~\cite{zhong2022adversarial} & 
     22.04\mypm0.36	& 30.83\mypm0.52 & 65.19\mypm0.82 &	64.96\mypm0.81 & 47.43\mypm0.72 & 31.90\mypm0.52 &	53.95\mypm0.79 & 35.81\mypm0.59	& \cellcolor{blue!15}44.01 \\

     & DSU~\cite{li2022uncertainty}& 
     22.35\mypm0.36 & 31.43\mypm0.51 & 64.55\mypm0.83 & 64.73\mypm0.81 & 47.74\mypm0.72 & 31.61\mypm0.53 & 54.81\mypm0.81 & 37.19\mypm0.61 & \cellcolor{blue!15}44.30 \\
     
     & \textbf{ Style-FGSM (ours)} &  
    \cellcolor{blue!15}
    \textbf{22.64\mypm0.35} &
	\cellcolor{blue!15}\textbf{33.96\mypm0.57} &
	\cellcolor{blue!15}\textbf{70.94\mypm0.82} &
	\cellcolor{blue!15}\textbf{74.13\mypm0.78} & 
	\cellcolor{blue!15}\textbf{48.49\mypm0.72} &
	\cellcolor{blue!15}\textbf{34.64\mypm0.57} &
	\cellcolor{blue!15}\textbf{58.58\mypm0.83} &
	\cellcolor{blue!15}\textbf{41.13\mypm0.67} & \cellcolor{blue!15}\textbf{48.06}\\
	
     \midrule
     5-shot 
      & StyleGaus$\dagger$ & 
     24.97\mypm0.37&
	41.74\mypm0.48&
	81.88\mypm0.61&
	89.71\mypm0.49&
	65.98\mypm0.67&
	45.03\mypm0.64&
	72.66\mypm0.68&
	56.66\mypm0.65& \cellcolor{blue!15}59.83
\\
       
     & MixStyle~\cite{zhou2021domain} & 
     25.04\mypm0.36&	
     43.77\mypm0.53&	
     82.67\mypm0.58&	
     88.90\mypm0.52&		
     65.73\mypm0.66&	
     45.91\mypm0.63&	
     75.90\mypm0.63&	
     56.59\mypm0.62& \cellcolor{blue!15}60.56 \\
     
     & AdvStyle~\cite{zhong2022adversarial} & 
     25.03\mypm0.35 & 43.15\mypm0.50 & 83.09\mypm0.57 & 88.44\mypm0.52 & 66.42\mypm0.67 & 44.85\mypm0.64 & 74.14\mypm0.65 & 54.89\mypm0.64 & \cellcolor{blue!15}60.00 \\

     & DSU~\cite{li2022uncertainty} & 25.02\mypm0.36 & 45.19\mypm0.52 & 80.30\mypm0.63 & 86.30\mypm0.56 & 67.94\mypm0.66 & 45.65\mypm0.63 & 75.17\mypm0.64 & 54.31\mypm0.62 & \cellcolor{blue!15}59.99\\

     & \textbf{Style-FGSM (ours)}&
     \cellcolor{blue!15}\textbf{26.07\mypm 0.37} &
    \cellcolor{blue!15}\textbf{45.77\mypm 0.51} &
	\cellcolor{blue!15}\textbf{86.58\mypm 0.54} &
	\cellcolor{blue!15}\textbf{93.65\mypm 0.39} & 
	\cellcolor{blue!15}\textbf{68.72\mypm 0.67} &
	\cellcolor{blue!15}\textbf{50.13\mypm 0.68} &
	\cellcolor{blue!15}\textbf{77.73\mypm 0.62} &
	\cellcolor{blue!15}\textbf{61.52\mypm 0.68} & \cellcolor{blue!15}\textbf{63.77}\\
     
	 \bottomrule   
 	\end{tabular}
    \end{adjustbox}
\vspace{-0.15in}
    \caption{\small
    Different style augmentation methods are compared. ``StyleGaus$\dagger$'' means adding random Gaussian noises to the styles, where $\dagger$ represents it is proposed by us.
    ``MixStyle~\cite{zhou2021domain}'', ``AdvStyle~\cite{zhong2022adversarial}'' and ``DSU~\cite{li2022uncertainty}'' are adapted from other tasks, e.g., domain generation. Results (\%) conducted under 5-way 1-shot/5-shot settings. Methods are built upon the ResNet10 and GNN.}
    \label{tab:StyleAug}
    \vspace{-0.2in}
\end{table*}

\noindent \textbf{Why Attack Styles Instead of Images or Features?}
A natural question may be why we choose to attack styles instead of other targets, e.g., the input image as in AQ~\cite{goldblum2020adversarially}, MDAT~\cite{li2019defensive}, and ATA~\cite{wang2021cross} or the features as in Shen et al.~\cite{shen2019learning} and AFA~\cite{hu2022adversarial}? To answer this question, we compare our StyleAdv which attacks styles against attacking images and features by modifying the attack targets of our method. The 5-way 1-shot/5-shot results are given in Table~\ref{tab:AttackObject}.
We highlight several points. 
1) We notice that attacking image, feature, and style all improve the base GNN model (given in Table~\ref{tab:main-result-new}) which shows that all of them boost the generalization ability of the model by adversarial attacks.
Interestingly, the results of our ``Attack Image''/``Attack Feature'' even outperform the well-designed CD-FSL methods ATA~\cite{wang2021cross} and AFA~\cite{hu2022adversarial} (shown in Table~\ref{tab:main-result-new});
2) Our method has clear advantages over attacking images and features. This again indicates the superiority of tackling visual styles for narrowing the domain gap issue for CD-FSL.

\noindent \textbf{Is Style-FGSM Better than Other Style Augmentation Methods?}
To show the advantages of our Style-FGSM against other style augmentation methods, we introduce several competitors including ``StyleGaus'', MixStyle~\cite{zhou2021domain}, AdvStyle~\cite{zhong2022adversarial}, and DSU~\cite{li2022uncertainty}. 
Typically, ``StyleGaus'' that adds random Gaussian noises into the styles is introduced as a simple but reasonable baseline. 
MixStyle~\cite{zhou2021domain}, AdvStyle~\cite{zhong2022adversarial}, and DSU~\cite{li2022uncertainty} which are initially designed for other tasks, e.g., segmentation and domain generation are also adapted.
The results are reported in Table~\ref{tab:StyleAug}.
Comparing the results of StyleGuas with that reported in Table~\ref{tab:main-result-new}, we find that perturbing the styles on the feature level by simply adding random noises also improves the base GNN and even surpasses a few CD-FSL competitors on some target datasets. This phenomenon is consistent with the insight that augmenting the style distributions helps boost the CD-FSL methods. 
As for the comparison between our Style-FGSM and other advanced style augmentation competitors, we find that Style-FGSM performs better than all the MixStyle, AdvStyle, and DSU on both 1-shot and 5-shot settings.
Typically, MixStyle and DSU both generate virtual styles, but their new styles are still relatively easy. This shows that our hard styles boost the model to a larger extent.
AdvStyle generates both virtual and hard (adversarial) styles. However, it is still inferior to us. This indicates the advantages of our method that attacks in latent feature space and adopts two individual tasks for attacking and optimization.

\section{Conclusion}
This paper presents a novel model-agnostic StyleAdv for CD-FSL. Critically, to narrow the domain gap which is typically in the form of visual shifts,  StyleAdv solves the minimax game of style adversarial learning: first adds perturbations to the source styles increasing the loss of the current model, then optimizes the model by forcing it to recognize both the clean and style perturbed data. 
Besides, a novel progressive style adversarial attack method termed style-FGSM is presented by us. Style-FGSM synthesizes diverse ``hard'' and ``virtual'' styles via adding the signed gradients to original clean styles. These generated styles support the max step of StyleAdv. Intuitively, by exposing the CD-FSL to adversarial styles which are more challenging than those limited real styles that exist in the source dataset, the generalization ability of the model is boosted. Our StyleAdv improves both CNN-based and ViT-based models. 
Extensive experiments indicate that our StyleAdv build new SOTAs.

\noindent\textbf{Acknowledgement.} This project was supported by National Key R\&D Program of China (No. 2021ZD0112804) and NSFC under Grant No. 62076067.

{\small
\bibliographystyle{ieee_fullname}
\bibliography{references}
}

\appendix

\twocolumn[
\begin{@twocolumnfalse}
\section*{\centering{Supplementary Material for \\ Meta Style Adversarial Training for Cross-Domain Few-Shot Learning}}
\end{@twocolumnfalse}
]

We first provide more implementation details in Sec.~\ref{sec:detail}; then we show more experimental results including plugging StyleAdv into different FSL/CD-FSL methods, building StyleAdv upon the PGD attacker, optimizing the model using different losses, 
and more ablation studies in Sec.~\ref{sec:exp};
Finally, in Sec.~\ref{sec:analyse}, we provide more visualization results.

\section{More Implementation Details}\label{sec:detail}
\subsection{Progressive Attacking Method}
\begin{figure}[h]
  \centering
  \vspace{-0.1in}
  \includegraphics[width=1
  \linewidth]{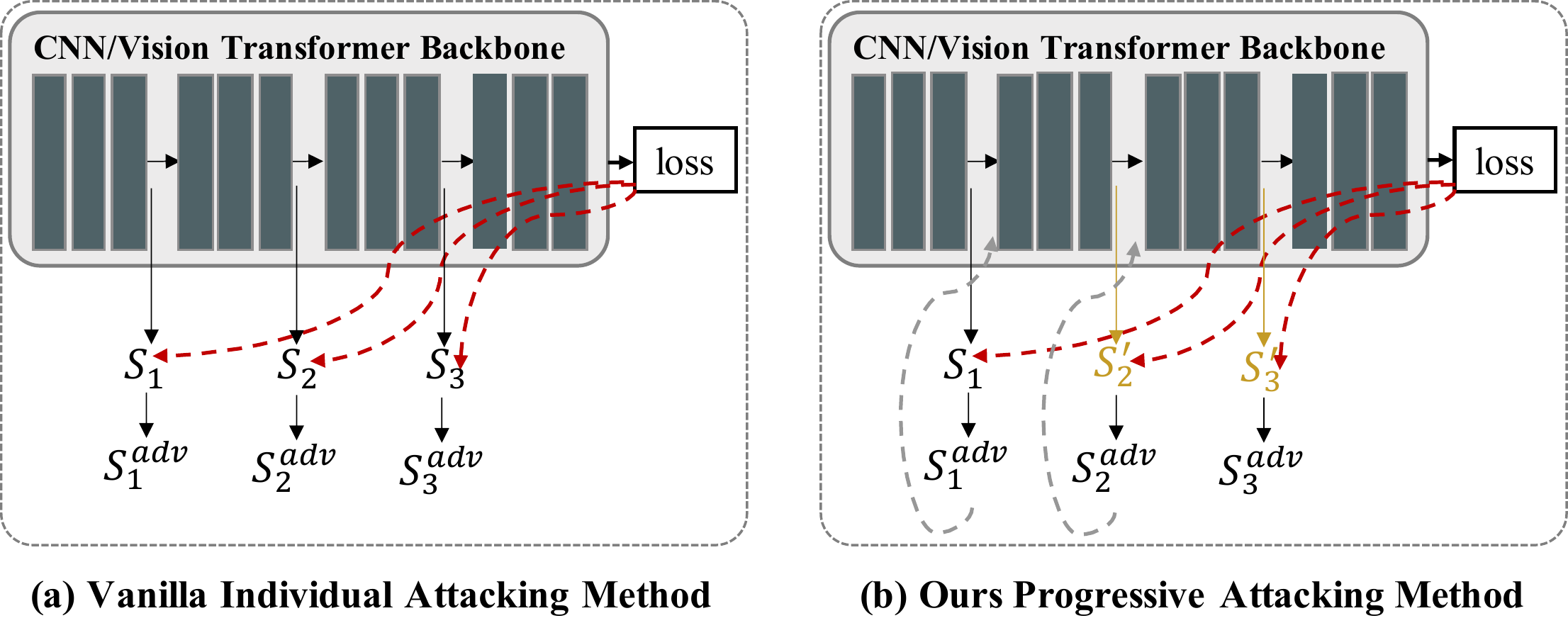}
   \vspace{-0.25in}
  \caption{Illustrations of the vanilla/progressive attacking methods.}
    \vspace{-0.2in}
  \label{fig:abla-progressive}
\end{figure}

To better help understand our proposed progressive attacking strategy, we compare it with the vanilla individual attacking approach. The illustrations are provided in Figure~\ref{fig:abla-progressive}. For simplification, we use $S_1$, $S_2$, and $S_3$ to represent the styles extracted from blocks $E_1$, $E_2$, and $E_3$, respectively. Correspondingly, $S_1^{adv}$, $S_2^{adv}$, and $S_3^{adv}$ represent the adversarial styles.

We would like to highlight two points:
1. The vanilla individual attacking method takes each block separately, which may lead to inconsistencies between features in different blocks.
2. By contrast, our progressive attacking method accumulates the adversarial signals, generating smooth adversarial features.
Overall, we take the dependencies between blocks into account and produce a more coherent set of adversarial features via the progressive attacking way.

\subsection{Loss Functions}
Given the clean and perturbed episode features $F_{\mathcal{T}}$ and $F_{\mathcal{T}}^{adv}$, recall that StyleAdv contains four sub losses: the global classification loss $\mathcal{L}_{cls}$, the original FSL loss $\mathcal{L}_{fsl}$, the adversarial FSL loss $\mathcal{L}_{fsl}^{adv}$, and the consistency loss $\mathcal{L}_{cons}$.

\noindent\textbf{Global Classification Loss}: The $\mathcal{L}_{cls}$ is the cross entropy (CE) loss between the predictions of the global classification scores $f_{cls}(F_{\mathcal{T}})$ and the global class labels $Y$. 

\noindent\textbf{Original/Adversarial FSL Loss}: Instead of using the global labels $Y$, meta-learning adopts local FSL class labels $Y_{fsl}$ for query images by adjusting the global labels to the set of $[0, 1, 2, ..., N-1]$, where N denotes the N classes contained in the episode. Since we perturb the episode at style level while maintain the semantic content unchanged, the synthesized adversarial data still belong to the same FSL label $Y_{fsl}$. The FSL losses thus are calculated as $\mathcal{L}_{fsl} = CE(P_{fsl}, Y_{fsl})$, $\mathcal{L}_{fsl}^{adv} = CE(P_{fsl}^{adv}, Y_{fsl})$, where $P_{fsl} = f_{fsl}(F_{\mathcal{T}})$, $P_{fsl}^{adv} = f_{fsl}(F_{\mathcal{T}}^{adv})$.

\noindent\textbf{Consistency Loss:} The $\mathcal{L}_{cons}$ is introduced to constrain the consistency between the prediction $P_{fsl}$ and $P_{fsl}^{adv}$. Specifically, it is calculated by the KL divergence loss which is defined below:
\begin{equation}
    \mathcal{L}_{cons} = \frac{1}{BN}\sum_{i=1}^B\sum_{j=1}^N P_{{fsl}_{(i,j)}}(\log P_{{fsl}_{(i,j)}} - \log P_{{fsl}_{(i,j)}}^{adv}),
\end{equation}
where $P_{fsl}, P_{fsl}^{adv} \in \mathcal{R}^{NM\times N}, B=NM$.

\subsection{Competitors}
In this paper, besides the existing CD-FSL methods, totally six competitors including ``Attack Image'', ``Attack Feature'' (as in Table~\ref{tab:AttackObject}) and ``StyleGaus'', ``MixStyle'', ``AdvStyle", and ``DSU'' (as in Table~\ref{tab:StyleAug}) are adapted. Thus, we give an introduction to the implementation details of these proposed competitors.

\noindent\textbf{Attack Image \& Attack Feature}: 
Generally, the ``Attack Image'' and ``Attack Feature'' share the same forward pipeline as our StyleAdv. To summarize, given a clean episode, all these three methods first perturb the original data via adversarial attack and then optimize the whole network under the supervision of both clean and adversarial perturbed episodes. The loss defined in Eq.~\ref{eq: train-loss} which contains four sub losses is utilized to optimize the network. Besides, the hyper-parameters are also kept consistent.
While different from attacking styles as in StyleAdv, ``Attack Image'' attacks the episode at the image pixel level. That is, given clean episode $(\mathcal{T}, Y)$, the attacked $\mathcal{T}^{adv}$ is defined as,

\begin{equation}\label{eq:attack-image}
\mathcal{T}^{adv} = \mathcal{T} + k_{RT}\cdot\mathcal{N}(0,I) + \epsilon \cdot sign(\nabla_{\mathcal{T}}J(\theta_{E}, \theta_{cls}, \mathcal{T}, Y)).
\end{equation}

While ``Attack Feature'' generates the adversarial feature $F_{\mathcal{T}^{adv}}$   from the clean episode feature $F_{\mathcal{T}}$ as,

\begin{equation}\label{eq:attack-feature}F_{\mathcal{T}}^{adv} = F_{\mathcal{T}} + k_{RT}\cdot\mathcal{N}(0,I) + \epsilon \cdot sign(\nabla_{F_{\mathcal{T}}}J(\theta_{E}, \theta_{cls}, F_{\mathcal{T}}, Y)).
\end{equation}

To ensure a more fair comparison, the features of different blocks are attacked in the same progressive strategy as StyleAdv.
Concretely, using $F_{1}$ denotes the feature extracted by the first block $E_{1}$ i.e. $F_{1} = E_{1}(\mathcal{T})$. The $F_{1}^{adv}$ can be easily obtained as in Eq.~\ref{eq:attack-feature}.
However, for the subsequent block $E_{2}$, rather than obtaining $F_{2}$ as $E_{2}(F_{1})$, we have $F_{2}^{'} = E_{2}(F_{1}^{adv})$. Attacking $F_{2}^{'}$ results in the $F_{2}^{adv}$. Similarly, we obtain the $F_{3}^{adv}$, thus get the final feature $F_{\mathcal{T}}^{adv}$ as the result of applying max pooling into the $F_{3}^{adv}$.

\noindent\textbf{StyleGaus:}
The only difference between StyleGuas and StyleAdv lies in that rather than synthesizing new styles by adversarial attack as in Eq.~\ref{eq:styleAttack-mean} and Eq.~\ref{eq:styleAttack-std}, StyleGaus adds random Gaussian noises into the style $(\mu, \sigma)$ as below:
\begin{equation}
\mu^{adv} = \mu + k\cdot\mathcal{N}(0,I), \sigma^{adv} = \sigma + k\cdot\mathcal{N}(0,I),
\end{equation}
where $k$ is set as $\frac{16}{255}$. 
Note that all the other implement details e.g. network modules, pipeline, losses, and progressive augment manner are the same as StyleAdv.

\noindent\textbf{MixStyle:} The results of adapting MixStyle~\cite{zhou2021domain} for CD-FSL are introduced from wave-SAN~\cite{fu2022wave}. 
Typically, the MixStyle competitor is constructed by randomly sampling two episodes from the source training set and using the mixed style of these two episodes as the new style. 

\noindent\textbf{AdvStyle:} We implement the AdvStyle that attacks the style on images according to the pseudo codes provided in its paper. However, AdvStyle is initially proposed for segmentation, while we tackle the CD-FSL problem. Once we set the task as N-way K-shot, we could not take data of different sizes as input. Thus, rather than concating the original episode and the style-attacked episode as the input, we perform FSL tasks for these two episodes in parallel and use the sum of two FSL losses to optimize the network. For fair comparisons, the attacking ratio is set as $[0.008, 0.08,0.8]$.

\noindent\textbf{DSU:} The DSU is adapted into CD-FSL by replacing our style attacking method as their method -- modeling a Gaussian style distribution for each current batch of training data, and then randomly sample a new style from the Gaussian style. The core codes for modeling the style as uncertain Gaussian are provided by DSU.

\subsection{Details for Finetuning}
For each novel testing episode, as stated in Sec.~\ref{sec:experiments}, we generate pseudo training episodes and use them for finetuning the meta-trained model.
Empirically, the finetuning stage is sensitive to the learning rates and tuning iterations. Thus, we provide the specific finetuning details as in Table~\ref{tab:finetune-detail}.
Overall, compared to the ViT-small with large pretrained parameters as initialization, the ResNet-10 (RN10) trained purely on the single source dataset requires a bigger learning rate; compared to the 5-shot models, finetuning 1-shot models needs fewer training iterations.

\begin{table}[h!]
    \scriptsize
    \centering  
    \begin{tabular}{l|c|c|c|c|c} %
    \toprule
	 \textbf{Backbone}&\textbf{LargeP} &  \textbf{Task} & \textbf{Optimizer}  & \textbf{Iter} & \textbf{LR} \\ 
     \midrule
     RN10 & - & 5-way 5-shot & Adam & 50 & \{0, 0.001\} \\
     RN10 & - & 5-way 1-shot & Adam & 10 & \{0, 0.005\} \\
     \midrule
     ViT-small & DINO/IN1K & 5-way 5-shot & SGD & 50 & \{0, 5e-5\} \\
     ViT-small & DINO/IN1K & 5-way 1-shot & SGD & 20 & \{0, 5e-5\} \\
	 \bottomrule   
 	\end{tabular}
    \caption{The finetuning details for our ResNet10 (RN10) and ViT-small based models. The ``LargeP" denotes the large-scale pretrained model. The ``Iter" and the "LR" represent the tuning iterations and the learning rate, respectively.}
    \vspace{-0.15in}
    \label{tab:finetune-detail}
\end{table}

\section{More Experimental Results}\label{sec:exp}
\subsection{Working in A Plug-and-Play Manner.}
We highlight that our StyleAdv is complementary to other CD-FSL methods and can be used in a plug-and-play manner. To validate that, we show the results of plugin our StyleAdv into several different base models. The results are reported in Table.~\ref{tab:main-result}.

From the results, we draw the conclusion that our StyleAdv is model-agnostic and improves other FSL/CD-FSL methods effectively. 
Concretely, taking four different FSL/CD-FSL methods as base models, our StyleAdv promotes performance in most cases. 
Taking 5-way 1-shot as an example, we on average improve the RelationNet~\cite{sung2018learning}, the GNN~\cite{garcia2017few}, the FWT~\cite{tseng2020cross}, and the PMF~\cite{hu2022pushing} by 4.81\%, 4.51\%, 3.75\%, and 2.29\%, respectively. Similar improvements can be observed in 5-shot results.

\setlength{\tabcolsep}{3pt} 
\begin{table*}[!t]
   \scriptsize
    \centering 
    \begin{adjustbox}{width=\textwidth}
    \begin{tabular}{l|c|cccc|cccc|c} %
    \toprule
      \textbf{1-shot} & \textbf{Method} & 
      \texttt{\bf ChestX} & \texttt{\bf ISIC}& \texttt{\bf EuroSAT} & \texttt{\bf CropDisease} & \texttt{\bf Cub} &
      \texttt{\bf Cars}& \texttt{\bf Places} & \texttt{\bf Plantae} & {\textbf{Average}}\\
      \midrule
       
      RelationNet~\cite{sung2018learning}   
      & - &21.95\mypm0.20 & 30.53\mypm0.30 & 49.08\mypm0.40 & 53.58\mypm0.40 & 41.27\mypm0.40 & 30.09\mypm0.30 & 48.16\mypm0.50 & 31.23\mypm0.30 & 38.24 \\

      & + \textbf{StyleAdv } & \cellcolor{blue!15}22.39\mypm0.30&
	\cellcolor{blue!15}32.19\mypm0.46 &
	\cellcolor{blue!15}58.55\mypm0.66 &
	\cellcolor{blue!15}62.37\mypm0.68 &
	\cellcolor{blue!15}45.94\mypm0.59 &
	\cellcolor{blue!15}31.91\mypm0.48 & 
	\cellcolor{blue!15}53.06\mypm0.67 &
	\cellcolor{blue!15}38.02\mypm0.54 & \cellcolor{blue!15}43.05 (\textbf{4.81}$\uparrow$)
\\

      \midrule
      GNN~\cite{garcia2017few}
      & - & 22.00\mypm0.46  
      & 32.02\mypm0.66 & 63.69\mypm1.03 & 64.48\mypm1.08 & 45.69\mypm0.68   & 31.79\mypm0.51   & 53.10\mypm0.80   & 35.60\mypm0.56 & 43.55\\

      & + \textbf{StyleAdv} & 
    \cellcolor{blue!15}22.64\mypm0.35 &
	\cellcolor{blue!15}33.96\mypm0.57 &\cellcolor{blue!15}70.94\mypm0.82 &
	\cellcolor{blue!15}74.13\mypm0.78 & 
	\cellcolor{blue!15}48.49\mypm0.72 &
	\cellcolor{blue!15}34.64\mypm0.57 &
	\cellcolor{blue!15}58.58\mypm0.83 &
	\cellcolor{blue!15}41.13\mypm0.67 & \cellcolor{blue!15}48.06  (\textbf{4.51}$\uparrow$)\\

	\midrule
    FWT ~\cite{tseng2020cross}   
      & - & 22.04\mypm0.44 & 31.58\mypm0.67 & 62.36\mypm1.05 & 66.36\mypm1.04 & 47.47\mypm0.75   & 31.61\mypm0.53   & 55.77\mypm0.79   & 35.95\mypm0.58  & 44.14\\

      & + \textbf{StyleAdv} &\cellcolor{blue!15}22.91\mypm0.37 & 
	\cellcolor{blue!15}35.05\mypm0.56&
	\cellcolor{blue!15}68.03\mypm0.81&
	\cellcolor{blue!15}73.84\mypm0.78&
	\cellcolor{blue!15}48.68\mypm0.72&	
	\cellcolor{blue!15}34.88\mypm0.58&
	\cellcolor{blue!15}59.15\mypm0.84&	
	\cellcolor{blue!15}40.60\mypm0.66 & \cellcolor{blue!15}47.89 (\textbf{3.75}$\uparrow$)\\

     \midrule
    PMF$^\ast$~\cite{hu2022pushing} & - &
    21.73\mypm0.30 &	30.36\mypm0.36 &	70.74\mypm0.63 &	80.79\mypm0.62 &	78.13\mypm0.66 &	37.24\mypm0.57 &	71.11\mypm0.71 &	53.60\mypm0.66 &  55.46
    \\
    
    & + \textbf{StyleAdv}  & \cellcolor{blue!15}22.92\mypm0.32 &\cellcolor{blue!15}33.05\mypm0.44 &\cellcolor{blue!15}72.15\mypm0.65 &\cellcolor{blue!15}81.22\mypm0.61 &\cellcolor{blue!15}84.01\mypm0.58 &\cellcolor{blue!15}40.48\mypm0.57 &\cellcolor{blue!15}72.64\mypm0.67 &\cellcolor{blue!15}55.52\mypm0.66 &\cellcolor{blue!15}57.75 (\textbf{2.29}$\uparrow$)\\

	 \midrule
	 \midrule

	 \textbf{5-shot} & \textbf{Method} & 
     \texttt{\bf ChestX} & \texttt{\bf ISIC}& \texttt{\bf EuroSAT} & \texttt{\bf CropDisease} & \texttt{\bf Cub} & \texttt{\bf Cars}& \texttt{\bf Places} & \texttt{\bf Plantae} & {\textbf{Average}}\\
     \midrule
     RelationNet~\cite{sung2018learning}   
       & - & 24.07\mypm0.20 & 38.60\mypm0.30 & 65.56\mypm0.40 & 72.86\mypm0.40 & 56.77\mypm0.40 & 40.46\mypm0.40 & 64.25\mypm0.40 & 42.71\mypm0.30 & 50.66\\

    & + \textbf{StyleAdv } & \cellcolor{blue!15}25.38\mypm0.31 &\cellcolor{blue!15}42.99\mypm0.44&
	\cellcolor{blue!15}72.42\mypm0.56&
	\cellcolor{blue!15}80.70\mypm0.51&	
	\cellcolor{blue!15}63.94\mypm0.56&
	\cellcolor{blue!15}43.71\mypm0.57&
	\cellcolor{blue!15}69.55\mypm0.56&
	\cellcolor{blue!15}52.05\mypm0.54 & \cellcolor{blue!15}56.34 (\textbf{5.68}$\uparrow$)\\

     \midrule
     
     GNN~\cite{garcia2017few} 
     & - & 25.27\mypm0.46 &  43.94\mypm0.67 & 83.64\mypm0.77 & 87.96\mypm0.67 & 62.25\mypm0.65   & 44.28\mypm0.63   & 70.84\mypm0.65   & 52.53\mypm0.59 & 58.84 \\

     & + \textbf{StyleAdv} 
    & \cellcolor{blue!15}26.07\mypm 0.37 &\cellcolor{blue!15}45.77\mypm 0.51 &
	\cellcolor{blue!15}86.58\mypm 0.54 &
	\cellcolor{blue!15}93.65\mypm 0.39 & 
	\cellcolor{blue!15}68.72\mypm 0.67 &
	\cellcolor{blue!15}50.13\mypm 0.68 &
	\cellcolor{blue!15}77.73\mypm 0.62 &
	\cellcolor{blue!15}61.52\mypm 0.68 & \cellcolor{blue!15}63.77 (\textbf{4.93}$\uparrow$) \\

    \midrule
     FWT ~\cite{tseng2020cross}   
       & - & 25.18\mypm0.45  & 43.17\mypm0.70 & 83.01\mypm0.79 & 87.11\mypm0.67 & 66.98\mypm0.68 & 44.90\mypm0.64 & 73.94\mypm0.67 & 53.85\mypm0.62 & 59.77\\

       & + \textbf{StyleAdv} &\cellcolor{blue!15}25.53\mypm0.36&
	\cellcolor{blue!15}{47.36\mypm0.53}&
	\cellcolor{blue!15}{85.74\mypm0.55}&
	\cellcolor{blue!15}{92.32\mypm0.45}&
	\cellcolor{blue!15}{70.25\mypm0.68}&
	\cellcolor{blue!15}{49.97\mypm0.66}&
	\cellcolor{blue!15}{78.78\mypm0.60}&
	\cellcolor{blue!15}{60.23\mypm0.65}&\cellcolor{blue!15}{63.77} (\textbf{4.00}$\uparrow$)\\

    \midrule

    PMF~\cite{hu2022pushing} & -  & 
     \cellcolor{blue!15}{27.27} 
    & \cellcolor{blue!15}50.12  
    & 85.98  
    & 92.96 & - & - & - & - & - \\
    
    & + \textbf{StyleAdv} &
   26.97\mypm0.33 &47.73\mypm0.44 &\cellcolor{blue!15}{88.57\mypm0.34} &\cellcolor{blue!15}{94.85\mypm0.31} &\cellcolor{blue!15}{95.82\mypm0.27} &\cellcolor{blue!15}{61.73\mypm0.62} &\cellcolor{blue!15}{88.33\mypm0.40} &\cellcolor{blue!15}{75.55\mypm0.54} &\cellcolor{blue!15}{72.44}\\

	\bottomrule   
 	\end{tabular}
    \end{adjustbox}
    \vspace{-0.1in}
    \caption{\textbf{Results of our StyleAdv working in a plug-and-play way.} Methods trained on mini-Imagenet and evaluated in eight various novel target datasets, respectively. ``-'' represents the base model, ``+StyleAdv'' means that our StyleAdv is applied to the base model. Results marked in blue perform best (best viewed in color).} \label{tab:main-result}
\end{table*}

\subsection{Working with Different Attack Algorithms?}

As stated in Sec.~\ref{sec:styleAttackMethod}, our style adversarial attack method is built upon the FGSM algorithm, thus we may wonder whether StyleAdv can still work with different attack algorithms.
To that end, we further propose a variant style attack method (Style-PGD) by adapting the PGD algorithm. Formally, 
\begin{equation}\label{eq:styleAttackPGD-RT}
\mu_{0}^{adv}  = \mu + k_{RT} \cdot \mathcal{N}(0,I), \sigma_{0}^{adv}  = \sigma + k_{RT} \cdot \mathcal{N}(0,I),
\end{equation}
\begin{equation}\label{eq:styleAttackPGD-mean}
\mu_{t}^{adv}  = \mu_{t-1}^{adv} + \epsilon \cdot
\mathrm{sign}(\nabla_{\mu}J(\theta_{E}, \theta_{f_{cls}}, \mathcal{A}(F_{\mathcal{T}}, \mu, \sigma), Y)),
\end{equation}
\begin{equation}\label{eq:styleAttackPGD-std}
\sigma_{t}^{adv}  = \sigma_{t-1}^{adv} + \epsilon \cdot
\mathrm{sign}(\nabla_{\sigma}J(\theta_{E}, \theta_{f_{cls}}, \mathcal{A}(F_{\mathcal{T}}, \mu, \sigma), Y)).
\end{equation}

The comparison results of the base GNN model, Style-PGD, and Style-FGSM are given in Table~\ref{tab:AttackAlg}.
Results show that both Style-PGD and Style-FGSM have a performance improvement against the base GNN. This basically shows that our StyleAdv is not sensitive to different attack algorithms. 
Besides, we also observe that Style-PGD is worse than Style-FGSM. This shows that the one-step attack is enough and more suitable to generate desired adversarial noises. Multi-step attacking may cause the generated styles too difficult to train the model. Besides, this significantly increases the burden of training. Thus, in this paper, we stick to the one-step Style-FGSM as our attack method.

\setlength{\tabcolsep}{3pt}
\begin{table*}[!h]
   \scriptsize
    \centering  
    \begin{adjustbox}{width=\textwidth}
    \begin{tabular}{l|c|cccc|cccc|c} %
    \toprule
	 & \textbf{Attack Algorithm} & 
     \texttt{\bf ChestX} & \texttt{\bf ISIC}& \texttt{\bf EuroSAT} & \texttt{\bf CropDisease} & \texttt{\bf Cub} & \texttt{\bf Cars}& \texttt{\bf Places} & \texttt{\bf Plantae} & {\textbf{Average}}\\
     \midrule
     1-shot  & 
     GNN~\cite{garcia2017few}  & 22.00\mypm0.46  & 32.02\mypm0.66 & 63.69\mypm1.03 & 64.48\mypm1.08 & 45.69\mypm0.68   & 31.79\mypm0.51   & 53.10\mypm0.80   & 35.60\mypm0.56 & 43.55\\ 
       
     & StyleAdv (Style-PGD) & 
     \cellcolor{blue!15}22.74\mypm0.35&	
     32.79\mypm0.53&	
     68.08\mypm0.82&	
     73.02\mypm0.81&		
     47.86\mypm0.70&	
     34.27\mypm0.56&	
     57.13\mypm0.83&	
     39.90\mypm0.63 & 46.97\\ 
     
     & \textbf{StyleAdv (Style-FGSM)} &  
     22.64\mypm0.35 &
	\cellcolor{blue!15}33.96\mypm0.57 &
	\cellcolor{blue!15}70.94\mypm0.82 &
	\cellcolor{blue!15}74.13\mypm0.78 & 
	\cellcolor{blue!15}48.49\mypm0.72 &
	\cellcolor{blue!15}34.64\mypm0.57 &
	\cellcolor{blue!15}58.58\mypm0.83 &
	\cellcolor{blue!15}41.13\mypm0.67 & \cellcolor{blue!15}48.06\\
     
     \midrule
     5-shot      &
     GNN~\cite{garcia2017few} & 25.27\mypm0.46 &  43.94\mypm0.67 & 83.64\mypm0.77 & 87.96\mypm0.67 & 62.25\mypm0.65   & 44.28\mypm0.63   & 70.84\mypm0.65   & 52.53\mypm0.59 & 58.84 \\ 
     & StyleAdv (Style-PGD) & 
     25.98\mypm0.38&	
     44.49\mypm0.50&	
     84.39\mypm0.57&	
     92.30\mypm0.43&		
     68.50\mypm0.67&	
     48.82\mypm0.64&	
     \cellcolor{blue!15}77.76\mypm0.62&	
     59.62\mypm0.66 & 62.73\\
     
     & \textbf{StyleAdv (Style-FGSM)}&
     \cellcolor{blue!15}26.07\mypm 0.37 &
    \cellcolor{blue!15}45.77\mypm 0.51 &
	\cellcolor{blue!15}86.58\mypm 0.54 &
	\cellcolor{blue!15}93.65\mypm 0.39 & 
	\cellcolor{blue!15}68.72\mypm 0.67 &
	\cellcolor{blue!15}50.13\mypm 0.68 &
	77.73\mypm 0.62 &
	\cellcolor{blue!15}61.52\mypm 0.68 & \cellcolor{blue!15}63.77\\
     
	 \bottomrule   
 	\end{tabular}
    \end{adjustbox}
     \vspace{-0.1in}
    \caption{\textbf{Results of StyleAdv working with different style attack algorithms.} Models are built upon the ResNet-10 and GNN.}
    \label{tab:AttackAlg}
\end{table*}

\subsection{Effectiveness of Each Loss Item.}
To show the effectiveness of each item, we conduct ablation studies on different losses. Concretely, we compare our StyleAdv which is optimized by four sub losses with that of ``w/o $\mathcal{L}_{cls}$'',  ``w/o $\mathcal{L}_{cons}$'', ``w/o $\mathcal{L}_{fsl}, \mathcal{L}_{cons}$'', and ``w/o $\mathcal{L}_{fsl}^{adv}, \mathcal{L}_{cons}$''. The 5-way 1-shot results are given in Table~\ref{tab:abla-loss}.

We first notice that all these variants perform worse than our method. This generally shows that each loss helps.
More specifically, 
comparing ``all losses'' with ``w/o $\mathcal{L}_{cls}$'', we observe that a obvious performance improvement is brought by $\mathcal{L}_{cls}$. It is not difficult to understand since the $\mathcal{L}_{cls}$ makes the global classifier optimized thus providing the correct gradients for the Style-FGSM. 
Also, by comparing the results of ours against that of ``w/o $\mathcal{L}_{cons}$'', we show that the consistency loss also contributes. It helps alleviate the semantic drift problem caused by perturbing the styles thus promoting the final model.
In addition, through the results of removing the $\mathcal{L}_{fsl}^{adv}$ and $\mathcal{L}_{cons}$, the effectiveness of the adversarial styles generated by us is well indicated. The model performance is boosted by 
introducing such relatively challenging styles.
Finally, we find that the original styles also help through the experimental results of ``w/o $\mathcal{L}_{fsl}$, $\mathcal{L}_{cons}$''.

\setlength{\tabcolsep}{3pt}
\begin{table*}[!h]
   \scriptsize
    \centering  
    \begin{adjustbox}{width=\textwidth}
    \begin{tabular}{l|l|cccc|cccc|c} %
    \toprule
	 & \textbf{Losses} & 
     \texttt{\bf ChestX} & \texttt{\bf ISIC}& \texttt{\bf EuroSAT} & \texttt{\bf CropDisease} & \texttt{\bf Cub} & \texttt{\bf Cars}& \texttt{\bf Places} & \texttt{\bf Plantae} & {\textbf{Average}}\\
     \midrule
     1-shot 
    & w/o $\mathcal{L}_{cls}$  & 
    22.36\mypm0.36&
	\cellcolor{blue!15}34.43\mypm0.57&
	67.86\mypm0.83&
	68.46\mypm0.80&
	48.13\mypm0.73&
	32.98\mypm0.56&
	56.44\mypm0.81&
	38.48\mypm0.63& 46.14 \\ 
	
	 & w/o $\mathcal{L}_{cons}$ & 
    \cellcolor{blue!15}22.68\mypm0.36&
	33.10\mypm0.53&
	70.06\mypm0.84&
	72.46\mypm0.80&
	48.34\mypm0.71&
	33.58\mypm0.55&
	57.65\mypm0.82&
	40.06\mypm0.64& 47.24\\

	& w/o $\mathcal{L}_{fsl}^{adv}, \mathcal{L}_{cons}$ & 22.05\mypm0.35&
	 32.49\mypm0.53&
	 68.86\mypm0.83&
	68.93\mypm0.81&
	47.23\mypm0.72&
	32.85\mypm0.57&
	55.88\mypm0.82&
	37.68\mypm0.62& 45.75 
 \\
		
	& w/o $\mathcal{L}_{fsl}, \mathcal{L}_{cons}$ &  22.34\mypm0.33&
	 34.29\mypm0.56&
	67.09\mypm0.82&
	73.23\mypm0.79&
	46.64\mypm0.69&
	\cellcolor{blue!15}35.10\mypm0.59&
	55.61\mypm0.79&
	40.44\mypm0.66& 46.84
 \\

    & All losses (ours) &  
     22.64\mypm0.35 &
	 33.96\mypm0.57 &
	\cellcolor{blue!15}70.94\mypm0.82 &
	\cellcolor{blue!15}74.13\mypm0.78 & 
	\cellcolor{blue!15}48.49\mypm0.72 &
	34.64\mypm0.57 &
	\cellcolor{blue!15}58.58\mypm0.83 &
	\cellcolor{blue!15}41.13\mypm0.67 & \cellcolor{blue!15}48.06\\
	 \bottomrule   
 	\end{tabular}
    \end{adjustbox}
    \caption{\textbf{Effectiveness of each loss item.} Results conducted under 5-way 1-shot setting. Models are built upon the ResNet-10 and GNN.}
    \label{tab:abla-loss}
\end{table*}

\setlength{\tabcolsep}{3pt}
\begin{table*}[!htbp]
   \scriptsize
    \centering  
    \begin{adjustbox}{width=\textwidth}
    \begin{tabular}{l|l|cccc|cccc|c} %
    \toprule
	 \textbf{1-shot}& \textbf{Choice} & 
     \texttt{\bf ChestX} & \texttt{\bf ISIC}& \texttt{\bf EuroSAT} & \texttt{\bf CropDisease} & \texttt{\bf Cub} & \texttt{\bf Cars}& \texttt{\bf Places} & \texttt{\bf Plantae} & {\textbf{Average}}\\
     \midrule
     RT
     & \XSolidBrush & 
    \cellcolor{blue!15}22.88\mypm0.35&
	33.93\mypm0.55&
	68.27\mypm0.82&
	72.40\mypm0.80&
	\cellcolor{blue!15}48.95\mypm0.70&
	\cellcolor{blue!15}35.36\mypm0.59&
	58.48\mypm0.81&	
	40.86\mypm0.66& 47.64
\\ 
     
     & \textbf{\Checkmark (ours)} &  
     22.64\mypm0.35 &
	\cellcolor{blue!15}33.96\mypm0.57 &
	\cellcolor{blue!15}70.94\mypm0.82 &
	\cellcolor{blue!15}74.13\mypm0.78 & 
    48.49\mypm0.72 &
	34.64\mypm0.57 &
	\cellcolor{blue!15}58.58\mypm0.83 &
	\cellcolor{blue!15}41.13\mypm0.67 & \cellcolor{blue!15}48.06\\
    \midrule

$p_{skip}$ & $p_{skip} = 0$ &
    22.59\mypm0.36&
	33.06\mypm0.52&
	67.26\mypm0.81&
	72.73\mypm0.79&		
	48.11\mypm0.70&
	\cellcolor{blue!15}35.92\mypm0.59&
	\cellcolor{blue!15}58.65\mypm0.82&
	40.43\mypm0.65& 47.34 
\\ 
     
     & $p_{skip} = 0.2$ &
     \cellcolor{blue!15}22.97\mypm0.37&	
     33.63\mypm0.54&
	70.06\mypm0.81&
	73.85\mypm0.78&
	48.06\mypm0.71&
	34.57\mypm0.58&
	58.43\mypm0.82&
	39.87\mypm0.65& 47.68
\\ 
     
     & \textbf{$p_{skip} = 0.4$ (ours)} &  
     22.64\mypm0.35 &
	33.96\mypm0.57 &
	\cellcolor{blue!15}70.94\mypm0.82 &
	\cellcolor{blue!15}74.13\mypm0.78 & 
	48.49\mypm0.72 &
	34.64\mypm0.57 &
	58.58\mypm0.83 &
	\cellcolor{blue!15}41.13\mypm0.67 & \cellcolor{blue!15}48.06\\
	
	 &$p_{skip} = 0.6$ &
	 22.54\mypm0.35&	
	 \cellcolor{blue!15}34.03\mypm0.55&
	 70.09\mypm0.81&
	 73.35\mypm0.80&
     \cellcolor{blue!15}48.68\mypm0.72&
	 33.78\mypm0.55&
	 58.28\mypm0.83&
	 40.24\mypm0.64& 47.62 \\
	 \midrule
	 
    $\epsilon_{list}$ & 
    $\epsilon_{list} = [20]$ &  20.83\mypm0.28 & 23.97\mypm0.34 & 50.68\mypm0.79 & 43.12\mypm0.73 & 29.41\mypm0.50 & 23.34\mypm0.35 & 32.79\mypm0.55 & 25.98\mypm0.41 & 31.27  \\
     & $\epsilon_{list} = [4]$  & 21.55\mypm0.32 & 29.06\mypm0.46 & 62.15\mypm0.78 & 61.56\mypm0.82 & 33.41\mypm0.56 & 28.55\mypm0.44 & 41.69\mypm0.67 & 32.77\mypm0.54 & 38.83  \\
    &$\epsilon_{list} = [1.6,0.16,0.016]$ &  
    \cellcolor{blue!15}22.71\mypm0.36&
	33.37\mypm0.54&
	\cellcolor{blue!15}70.98\mypm0.82&
	73.33\mypm0.79&
	\cellcolor{blue!15}48.76\mypm0.72&
	\cellcolor{blue!15}35.34\mypm0.60&
	58.25\mypm0.81&
	41.00\mypm0.65& 47.97
\\
    
    & \textbf{$\epsilon_{list} = [0.8,0.08,0.008]$ (ours)} &  
    22.64\mypm0.35 &
	\cellcolor{blue!15}33.96\mypm0.57 &
	70.94\mypm0.82 &
	\cellcolor{blue!15}74.13\mypm0.78 & 
	48.49\mypm0.72 &
	34.64\mypm0.57 &
	\cellcolor{blue!15}58.58\mypm0.83 &
	\cellcolor{blue!15}41.13\mypm0.67 & \cellcolor{blue!15}48.06 \\
	
     & $\epsilon_{list} = [0.4,0.04,0.004]$ & 
    22.66\mypm0.36&
	33.24\mypm0.53&
	69.10\mypm0.80&
	72.97\mypm0.79&
	48.21\mypm0.71&
	33.67\mypm0.57&
	57.58\mypm0.80&
	40.62\mypm0.66& 47.26 \\

     & $\epsilon_{list} = [0.2,0.02,0.002]$ & 
     22.47\mypm0.36&	
     32.30\mypm0.52&	
     68.22\mypm0.79&	
     72.06\mypm0.78&
	47.41\mypm0.71&
	33.60\mypm0.58&
	57.57\mypm0.83&	
	40.03\mypm0.64& 46.71 
 \\
	 \bottomrule   
 	\end{tabular}
    \end{adjustbox}
    \caption{\textbf{Ablation studies on the random start (RT), skip probability $p_{skip}$, and attacking ratio $\epsilon_{list}$.} 5-way 1-shot meta tasks are conducted. Models are built on ResNet10 and GNN.}
    \label{tab:abla-all}
\end{table*}

\subsection{More Ablation Studies of StyleAdv.}
Our StyleAdv perturbs the initial style using the attacking ratio randomly sampled from $\epsilon_{list}$ with a random skip probability $p_{skip}$. In addition, the operation of random start is applied before attacking.
Thus, we perform ablation studies on attacking with/without random start (RT), $p_{skip}$, and $\epsilon_{list}$. 
Specifically, for the random skip probability $p_{skip}$, we set it as 0, 0.2, 0.4 (ours), and 0.6, respectively.
For the attacking ratio $\epsilon_{list}$, four regular choices including $[0.2, 0.02, 0.002]$, $[0.4, 0.04, 0.004]$, $[0.8, 0.08, 0.008]$ (ours), and $[1.6, 0.16, 0.016]$ and two relative large options including $\epsilon_{list} = [4]$ and $\epsilon_{list} = [20]$ are conducted. The 5-way 1-shot results are given in Table~\ref{tab:abla-all}.

\textbf{1) With/without random start.}
We first notice that our choice of applying RT performs better than without RT in most cases with an average improvement of 0.42\%. 

\textbf{2) Different choices of $p_{skip}$.} For different choices of $p_{skip}$, we find that except for the Cars and Places, as $p_{skip}$ increases, the accuracy will first rise and then fail, or keep rising in some cases. This generally indicates that an appropriate $p_{skip}$ can trade off the introduced perturbations and the difficulty of the meta task. 

\begin{figure}[h!]
  \centering
  \includegraphics[width=1.0 \linewidth]{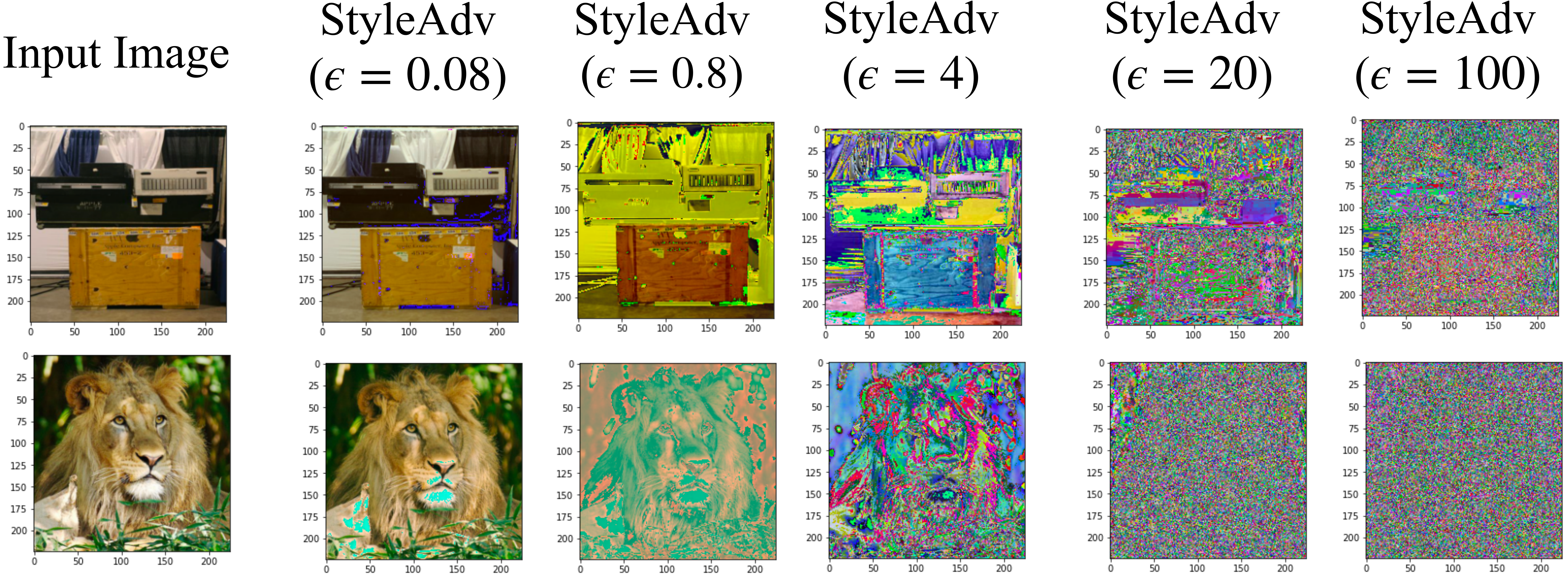}
  \vspace{-0.25in}
  \caption{Visualization results of stylized images generated by different attacking ratios.}
    \vspace{-0.15in}
  \label{fig:abla-largeStyle}
\end{figure}

\textbf{3) Different attacking ratios.} The phenomenons presented by different $\epsilon_{list}$ factually are basically similar to those of $p_{skip}$. The higher the value of $\epsilon_{list}$, the more difficult the meta task is. For the two large choices $\epsilon_{list} = [4]$ and $\epsilon_{list} = [20]$, we find that when the attacking ratio becomes too large, the perturbations added will affect the original semantic label, thus leading to the drastic performance drop. The visualization results of stylized images with large attacking ratios shown in Figure~\ref{fig:abla-largeStyle} further validate that a suitable attacking ratio is key.
In this paper, we set  $\epsilon_{list}$ as $[0.8, 0.08, 0.008]$ as a trade-off.
Note that our model is only trained with the mini-Imagenet without any single target image and we don't tune our model e.g. hyper-parameters for different target sets, thus it is unrealistic for our method to achieve totally consistent performance on eight unseen datasets. Alternatively, the choice with the relatively higher average performance is finally selected.

\begin{figure*}[h!]
  \centering
  \includegraphics[width=0.85 \linewidth]{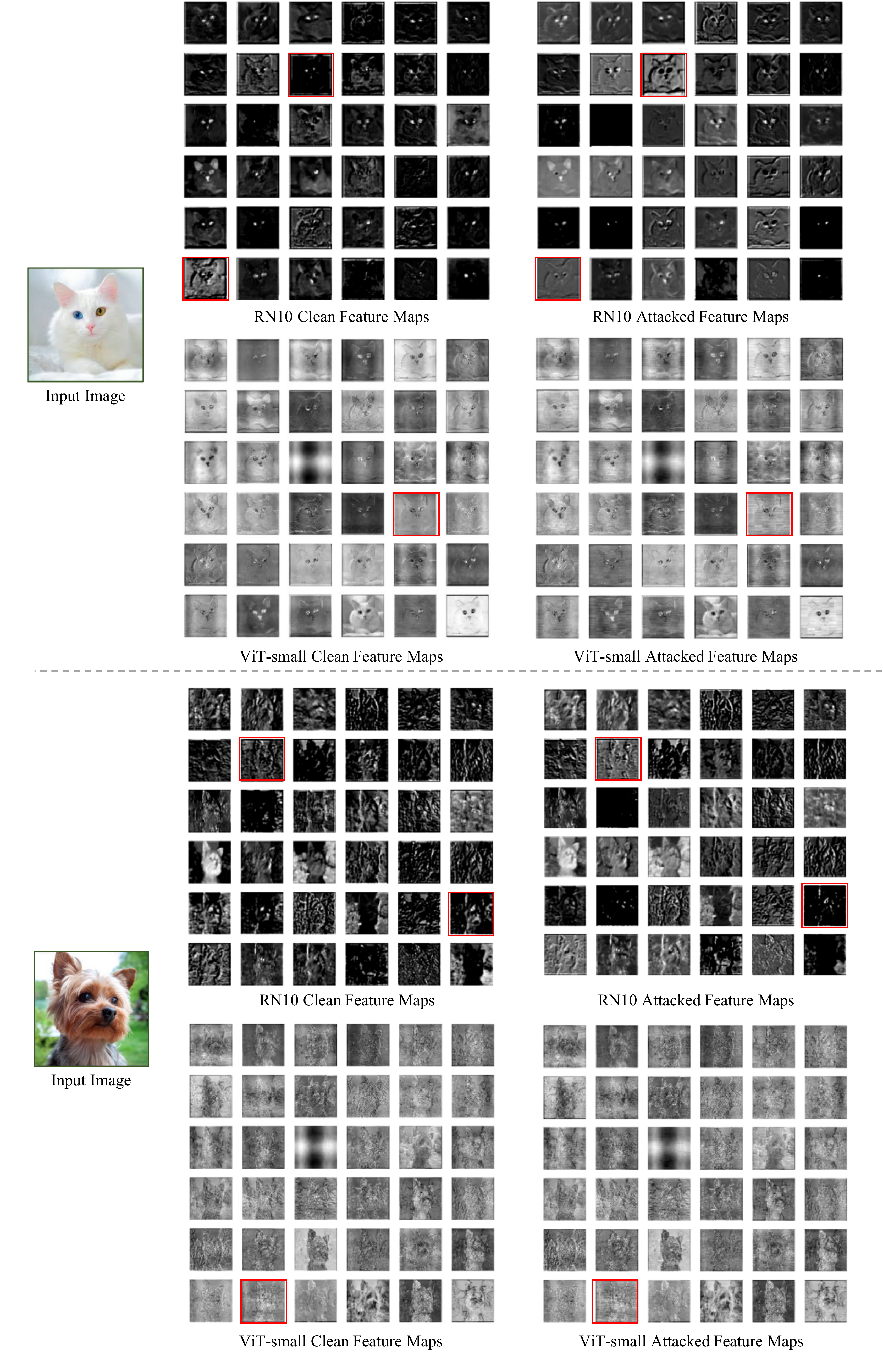}
  \vspace{-0.25in}
  \caption{The clean/style-attacked feature maps of ResNet10 (RN10) and ViT-small are visualized. We visualize 36 channels.}
  \vspace{-0.25in}
  \label{fig:more-vis}
\end{figure*}

\section{More Visualization Results.}\label{sec:analyse}
In the main file, as in Figure~\ref{fig:vis-new}, the stylized images generated by StyleAdv are given. Further, we provide the visualization results of feature maps extracted by both the ResNet-10 (RN10) and the ViT-small backbones. As shown in Figure~\ref{fig:more-vis}, two examples are illustrated. For each example, both the clean feature maps and the attacked feature maps are shown. The attacking ratio is set as $[0.008]$. Whether for RN10-based features or ViT-small-based features, we visualize 36 channels. Note that the VIT-small feature maps are formed by reshaping the patch tokens as we do in Sec.~\ref{sec:style}.

Results show that:
1) The ViT-small feature maps also correctly reflect the original image information e.g., the shapes. This validates our idea of the patch tokens still remain the spatial information and the whole image feature can be formed by reshaping the patch tokens. This further supports us to apply StyleAdv to the ViT features.
2) Since the visualization is performed on the gray feature maps, the differences between the clean feature maps and the attacked feature maps are somewhat not so significant. However, as highlighted in red bounding boxes, we can still observe minor changes.

\section{Discussion of Limitations}\label{sec:limitation}
As indicated in Table~\ref{tab:main-result-new}, wave-SAN outperforms StyleAdv on the Cub dataset. This result suggests that when the visual appearances of the source and target datasets are similar, augmenting the source styles via attacking may result in overly challenging meta-tasks. Although we still improve all the base models, exploring better methods to address this issue could be one of our further work.

\end{document}